\documentclass[10pt,twocolumn,letterpaper]{article}

\usepackage{cvpr}
\usepackage{fancyhdr}
\usepackage{times}
\usepackage{epsfig}
\usepackage{graphicx}
\usepackage{amsmath}
\usepackage{amsthm}
\usepackage{amssymb}
\usepackage{booktabs}
\usepackage{multirow}

\usepackage{tabularx}
\usepackage{cite}

\usepackage[safe]{tipa} %

\usepackage{algorithm}
\usepackage{algpseudocode}
\usepackage[frozencache,cachedir=.]{minted}
\usepackage{pifont}%
\newcommand{\xmark}{\ding{55}}%

\newcommand*{\R}{\mathbb{R}}
\newcommand*{\X}{\mathcal{X}}

\newcommand*{\Z}{\mathcal{Z}}

\newcommand*{\pr}{\mathbb{P}}
\newcommand*{\bfx}{\mathbf{x}}

\newcommand*{\bfz}{\mathbf{z}}

\DeclareMathOperator*{\argmin}{arg\,min}

\newtheorem{theorem}{Theorem}[section]

\usepackage[pagebackref=true,breaklinks=true,colorlinks,bookmarks=false]{hyperref}
\usepackage[capitalise, noabbrev]{cleveref}

\begin{document}

\title{Greedy-DiM: Greedy Algorithms for Unreasonably Effective Face Morphs}

\author{Zander W.~Blasingame\\
Clarkson University\\
Potsdam, NY, USA\\
{\tt\small blasinzw@clarkson.edu}\and
Chen Liu\\
Clarkson University\\
Potsdam, NY, USA\\
{\tt\small cliu@clarkson.edu}
}

\renewcommand{\headrulewidth}{0pt}
\renewcommand{\footrulewidth}{0.4pt}

\maketitle

\fancypagestyle{empty}{
    \fancyfoot[LO]{Accepted as a conference paper at IJCB 2024.}
}

\pagestyle{fancy}
\fancyhead{}
\fancyfoot[LO]{Accepted as a conference paper at IJCB 2024.}

\begin{abstract}
    Morphing attacks are an emerging threat to state-of-the-art Face Recognition (FR) systems, which aim to create a single image that contains the biometric information of multiple identities. Diffusion Morphs (DiM) are a recently proposed morphing attack that has achieved state-of-the-art performance for representation-based morphing attacks. However, none of the existing research on DiMs have leveraged the iterative nature of DiMs and left the DiM model as a black box, treating it no differently than one would a Generative Adversarial Network (GAN) or Varational AutoEncoder (VAE). We propose a greedy strategy on the iterative sampling process of DiM models which searches for an optimal step guided by an identity-based heuristic function. 
    We compare our proposed algorithm against ten other state-of-the-art morphing algorithms using the open-source SYN-MAD 2022 competition dataset. We find that our proposed algorithm is unreasonably effective, fooling all of the tested FR systems with an MMPMR of  100\%, outperforming all other morphing algorithms compared.
\end{abstract}

\section{Introduction}
\begin{figure}[t]
    \centering
    \begin{subfigure}{0.33\linewidth}
        \includegraphics[width=0.98\textwidth]{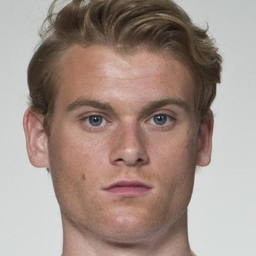}
        \caption{Identity $a$}
    \end{subfigure}%
    \begin{subfigure}{0.33\linewidth}
        \includegraphics[width=0.98\textwidth]{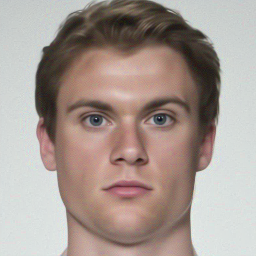}
        \caption{Morphed image}
    \end{subfigure}%
    \begin{subfigure}{0.33\linewidth}
        \includegraphics[width=0.98\textwidth]{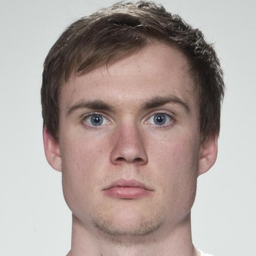}
        \caption{Identity $b$}
    \end{subfigure}
    \caption{Example of a morph created using Greedy-DiM. Samples are from the FRLL dataset~\cite{frll}.}
    \label{fig:morph_ex}
\end{figure}

Face Recognition (FR) systems have been deployed in a wide range of settings from unlocking phones to border control~\cite{frs-rates}, and the performance of such systems keeps improving. 
Due to its wide adoption, FR systems have been the target to various types of attacks~\cite{nist-frvt-morph}.
Most recently, it has been shown that these systems are vulnerable to the face morphing attacks~\cite{Ferrara2016,fraud_id,morphed_first,sebastien_on_detection_of_ma_gan,Blasingame2021LeveragingAL}, an emerging type of attack that aims to blend the biometric identifiers of multiple faces into a single image that would trigger a false acceptance with any of the identities used to make the morph, see~\cref{fig:morph_ex} for an illustration.
Application scenarios where the system allows users to submit self-generated images to obtain a valid passport or visa creates a significant attack vector for a malicious agent to enroll morphed images~\cite{busch_face_morph_survey}.
The potency of these attacks have led to face morphing becoming a focal point in recent research on FR systems~\cite{Ferrara2014TheMP, mad-data-eval,vgg19-mad,multi-fusion-smad,nasser_wavelet,Venkatesh_2020_WACV}, in an effort to secure the identity control process.

Traditional morph generation methods are based on pixel-level features, \eg, aligning facial landmarks, which could result in numerous artefacts~\cite{sebastien_gan_threaten}.
Recent work has focused on using generative AI models for representation-based attacks as the morphing occurs in the representation space rather than the image space.
Generative Adversarial Networks (GANs)~\cite{mipgan,sebastien_gan_threaten,can_gan_beat_landmark,sebastien_on_detection_of_ma_gan} can generate powerful morphs with less artefacts, when compared to the pixel-level morphs~\cite{can_gan_beat_landmark}.
However, diffusion models have come to supersede GANs as the state-of-the-art backbone for generative AI research~\cite{diff_beat_gan}.
Recent work has shown that \textbf{Di}ffusion \textbf{M}orphs (\textbf{DiM}) can achieve state-of-the-art morphing performance~\cite{blasingame_dim,fast_dim, morph_pipe}.
While GAN-based morphs blend the latent representations of two component identities, either by an element-wise average or by optimizing the blended latent with an identity-based loss~\cite{mipgan}, the latent representation of diffusion models is not inherently suited for this process for two reasons.
One, the latent space is a high dimensional space with the same size as the image space and two, the forward encoding process is stochastic and not deterministic.
To remedy this, the primary neural network backbone in the diffusion model is conditioned on a conditional embedding of the target image and a deterministic forward pass~\cite{diffae} is used to encode the target image.
These twin representations, the conditional embedding and the deterministic latent, are then used in the morphing process.
The current state-of-the-art methods simply blend the two latent representations, either with a fixed blend value~\cite{blasingame_dim,fast_dim} or a large batch of morphs with different blend values~\cite{morph_pipe}, with the best performing blend value selected out of the batch.

While GAN-based methods can directly optimize the latent representation used to produce the morphed image~\cite{mipgan},
the iterative generative process of diffusion models, often requiring many steps to sample a high quality sample~\cite{blasingame_dim}, does not immediately provide such an analog.
We propose to use a greedy strategy to solve this optimization problem by locally choosing the optimal solution at each timestep in the sampling process.
We call the family of algorithms which use the greedy strategy: Greedy-DiM. %
Greedy-DiM optimizes the diffusion process for face morphing by using a greedy search strategy in order to create high-quality morphs while retaining minimal computational overhead when compared to the original DiM approach~\cite{blasingame_dim} or less overhead than later work~\cite{morph_pipe}.
We summarize our contributions in this work as follows:
\begin{enumerate}
    \item We propose a novel family of face morphing algorithms, Greedy-DiM.
    \item We present thorough experimental and theoretical analysis of the proposed morphing algorithms.
    \item To the best of our knowledge Greedy-DiM is the first representation-based morphing algorithm that \textit{consistently} outperform landmark-based morphing algorithms. %
\end{enumerate}

\begin{table*}[t]
    \centering
    \caption{Comparison of existing DiM methods in the literature and our proposed algorithm.}
    \footnotesize
    \begin{tabularx}{\linewidth}{@{\extracolsep{\fill}}lllll}
    \toprule
    & \textbf{DiM~\cite{blasingame_dim}} & \textbf{Fast-DiM~\cite{fast_dim}} & \textbf{Morph-PIPE~\cite{morph_pipe}} & \textbf{Ours (Greedy-DiM)}\\
    \midrule
    ODE solver & DDIM & DPM++ 2M & DDIM & DDIM\\
    Forward ODE solver & DiffAE & DDIM & DiffAE & DiffAE\\
    Number of sampling steps & 100 & 50 & 2100 & 20\\
    Heuristic function & \xmark & \xmark & $\mathcal{L}_{ID}^*$ & $\mathcal{L}_{ID}^*$\\
    Search strategy & \xmark & \xmark & Brute-force search & Greedy optimization\\ 
    Search space & $\emptyset$ & $\emptyset$ & Set of 21 blend values & Image space\\
    Probability the search space contains the optimal solution & \xmark & \xmark & 0 & 1\\
    \bottomrule
    \end{tabularx}
    \label{tab:overview}
\end{table*}

\section{Prior Work}
The na\"ive approach to create a face morph is to simply take a pixel-wise average of the original images.
Due to its simplicity, this approach can result in many noticeable artifacts. 
An extension on this would be to align the landmarks of the face, warping the pixels to fit this new representation, and then averaging the pixels.
The blending value is usually set to $0.5$ so that both of the original images could contribute equally to the morphed image. Later work~\cite{mipgan,morph_pipe} questions if this assumption is valid.
Further improvements can be made to these landmark-based morphs by additional post-processing to remove the artefacts from the morphing process~\cite{ferrara_post_processing}.

Morphing through Identity Prior driven GAN (MIPGAN) and in particular MIPGAN-II improve upon the original GAN-based morphs~\cite{morgan} and StyleGAN2-based morphs~\cite{sebastien_gan_threaten} by using optimization methods to search for a latent representation that fools the FR system maximally~\cite{mipgan}.
The MIPGAN algorithm works by taking the two bona fide images $x_a, x_b$, and encoding them into latent representations, $z_a, z_b$.
An initial morphed latent $z_{ab} = \frac 12 (z_a + z_b)$ is then constructed.
The initial latent is optimized such that the generated morph $x_{ab} = G(z_{ab})$ minimizes an identity loss based on the ArcFace FR system~\cite{deng2019arcface}  along with a perceptual loss term.
This optimization procedure produces a more potent morphing attack than simply using the morphed latent $z_{ab}$.

\subsection{DiM}
Blasingame~\etal~\cite{blasingame_dim} proposed to use diffusion-based methods to construct high-quality face morphs. This family of morphing attacks are known as Diffusion Morphs (DiM).
Diffusion models function by modeling the diffusion process wherein noise is progressively added to the data until the resulting distribution reassembles a Gaussian distribution~\cite{song2021denoising, Karras2022edm}.
To sample the data distribution, a random sample is drawn from the Gaussian distribution and then the diffusion process is run in reverse to iteratively remove noise until a clean sample from data distribution is produced.
Given the data distribution $p_{data}(\bfx)$ on the data space $\X \subseteq \R^n$, the diffusion process is given by the It\^o stochastic differential equation (SDE)
\begin{equation}
    \mathrm{d}\bfx_t = f(t)\bfx_t\; \mathrm dt + g(t)\; \mathrm d\mathbf{w}_t
\end{equation}
where $t \in [0, T]$ is the time with fixed constant $T > 0$, $f(\cdot)$ and $g(\cdot)$ are the drift and diffusion coefficients, and $\mathbf{w}_t$ denotes standard Brownian motion.
In this paper we employ the variance preserving (VP) formulation of diffusion models~\cite{song2021scorebased} which uses the following drift and diffusion coefficients
\begin{align}
    f(t) &= \frac{\mathrm d \log \alpha_t}{\mathrm dt} \\
    g^2(t) &= \frac{\mathrm d \sigma_t^2}{\mathrm dt} - 2 \frac{\mathrm d \log \alpha_t}{\mathrm dt} \sigma_t^2
\end{align}
where $\alpha_t, \sigma_t$ form the noise schedule of the diffusion model such that
\begin{equation}
    q(\bfx_t \mid \bfx_0) = \mathcal{N}(\bfx_t \mid \alpha_t \bfx_0, \sigma_t^2 \mathbf{I})
\end{equation}
with a signal-to-noise-ratio (SNR) $\alpha_t^2/\sigma_t^2$, which is strictly decreasing with respect to $t$.

The distribution of $\bfx_t$ is denoted as $p_t(\bfx_t)$, therefore $p_0(\bfx_0) \equiv p_{data}(\bfx)$.
Song~\etal~\cite{song2021scorebased} showed the existence of an ordinary differential equation (ODE) dubbed the Probability Flow ODE (PF-ODE) whose marginal distributions $p_t$ follow the same trajectories as the diffusion SDE. The PF-ODE is given as
\begin{equation}
    \label{eq:pf-ode}
    \frac{\mathrm d \bfx_t}{\mathrm dt} = f(t)\bfx_t - \frac{1}{2}g^2(t) \nabla_\bfx \log p_t(\bfx_t)
\end{equation}
where $\nabla_\bfx \log p_t(\bfx_t)$ denotes the score of $p_t(\bfx_t)$.
The score function can be learned by a neural network $\boldsymbol\epsilon_\theta(\bfx_t, t) \approx -\sigma_t\nabla_\bfx \log p_t(\bfx_t)$.
To sample the model, an initial noise sample $\bfx_T \sim \mathcal{N}(\mathbf 0, \mathbf I)$ is drawn.
Then a numerical ODE solver is employed to solve the ODE from time $T$ to time $0$.

DiM models employ score predictor model conditioned on a latent representation of the target image, \ie, $\boldsymbol\epsilon_\theta(\bfx_t, \bfz, t)$ with an encoding network $\bfz = \mathcal{E}(\bfx_0)$.
Previous DiM approaches~\cite{blasingame_dim,morph_pipe,fast_dim} use a Diffusion Autoencoder model~\cite{diffae} as the backbone for the score prediction model.
As per the Diffusion Autoencoder technique, the ODE solver $\Phi$ is reversed now solving from $t=0$ to $t=T$, resulting in a deterministic forward pass to an approximation of $p_T(\bfx_T)$. We denote this forward ODE solver, so called as it is evaluated as time runs forwards from $0$ to $T$, as $\Phi^{+}$.
The noisy bona fide images are morphed using spherical linear interpolation, \ie, $\bfx_T^{(ab)} = \mathrm{slerp}(\bfx_0^{(a)}, \bfx_0^{(b)}; 0.5)$.
The conditionals are averaged and this morphed information is used as the input to the diffusion model.
The DiM generative pipeline is provided in~\cref{appendix:dim_alg}.

Zhang~\etal~\cite{morph_pipe} proposed Morph-PIPE, a simple extension on the DiM method by using an identity-based loss to select between a range of blend values.
They use an identity loss defined as the sum of two sub-losses:
\begin{align}
    \mathcal{L}_{ID} &= d(v_{ab}, v_a) + d(v_{ab}, v_b)\label{eq:loss_id_part}\\
    \mathcal{L}_{diff} &= \big|d(v_{ab}, v_a) - d(v_{ab}, v_b))\big |\label{eq:loss_id_diff}\\
    \mathcal{L}_{ID}^* &= \mathcal{L}_{ID} + \mathcal{L}_{diff}\label{eq:loss_id}
\end{align}
where $v_a = F(\bfx_0^{(a)}), v_b = F(\bfx_0^{(b)}), v_{ab} = F(\bfx_0^{(ab)})$, and $F: \X \to V$ is an FR system which embeds images into a vector space $V$ which is equipped with a measure of distance, $d$.
To create the morph $B$, different morphs are generated with different blend coefficients $\{w_n\}_{n=1}^B \subseteq [0, 1]$.
They then choose the $w_n$ which minimizes $\mathcal{L}_{ID}^*$.
By leveraging a powerful FR system like ArcFace, the selected morphs are more potent than the original DiM implementation with a fixed blend parameter $w = 0.5$.

Fast-DiM is a concurrent work which proposes an improvement on DiM with the goal of reducing the number of Network Function Evaluations (NFE) needed to produce high-quality face morphs using DiM~\cite{fast_dim}.
This reduction in NFE is primarily accomplished by using alternative ODE solvers used in solving the Probability Flow ODE and replacing the ``stochastic encoder'' with an ODE solver for the Probability Flow ODE as time runs forward from $0$ to $T$.

\section{Greedy-DiM}
We propose Greedy-DiM, a novel family of face morphing attacks that greedily choose the optimal solution at each timestep in
solving the PF-ODE.
In~\cref{tab:overview} we present a high-level overview of our proposed approach and how it differs from existing~\cite{blasingame_dim,morph_pipe} and concurrent~\cite{fast_dim} work.
We evaluate the proposed morphing attacks on the open-source SYN-MAD 2022 dataset~\cite{syn-mad22} with the ArcFace~\cite{deng2019arcface}, ElasticFace~\cite{elasticface}, and AdaFace~\cite{adaface} FR systems via the Mated Morph Presentation Match Rate (MMPMR) metric~\cite{mmpmr} and number of Network Function Evaluations (NFE).
For our heuristic function $\mathcal{H}$ we use the identity loss outlined in~\cref{eq:loss_id} with the ArcFace FR system.
Further details on the experimental setup are provided in~\cref{sec:experiemental_setup}.

\begin{figure*}[t]
    \centering
    \includegraphics[trim={3cm 0 3cm 0},clip,width=\textwidth]{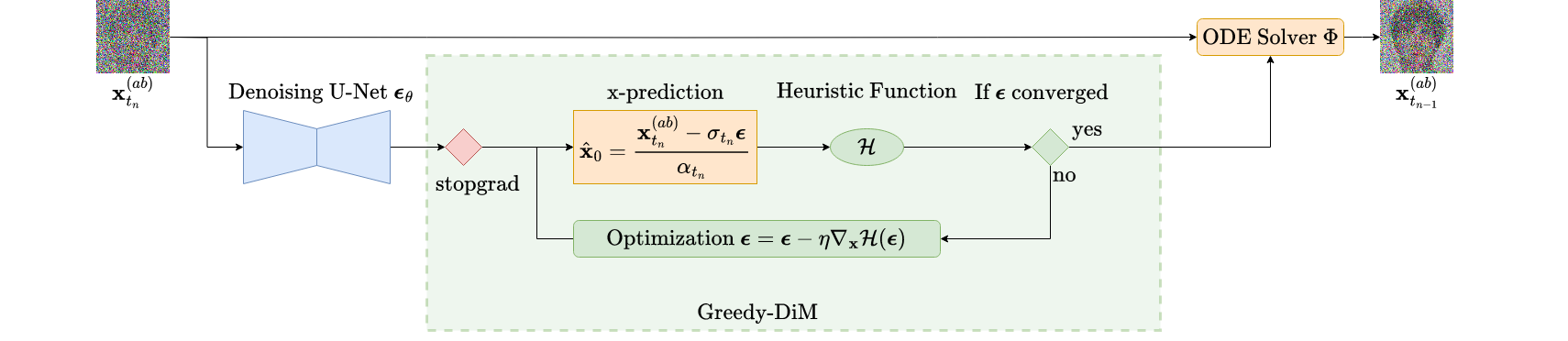}
    \caption{Overview of a single step of the Greedy-DiM* algorithm. Proposed changes highlighted in green.}
    \label{fig:greedy_dim_overview}
\end{figure*}

\subsection{Greedy Search}
We begin by applying a greedy search during the sampling process of diffusion models.
The information from the identity losses of~\cref{eq:loss_id,eq:loss_id_diff} provides a powerful heuristic function for guiding the creation of face morphs.
While Zhang~\etal~\cite{morph_pipe} only uses this information to select from a pre-selected set of morphed image candidates, we propose to use this heuristic during the sampling process.
Now, the identity losses require a morphed image as input; however, at time $t > 0$, the image $\bfx_t$ is not fully denoised.
To remedy this, we propose to use the estimate of $\bfx_0$ at each timestep $t$ as input to the identity losses. This can be found from the noise prediction network via
\begin{equation}
    \label{eq:x0_pred}
    \bfx_\theta(\bfx_t, \bfz, t) = \frac{\bfx_t - \sigma_t\boldsymbol{\epsilon}_\theta(\bfx_t, \bfz, t)}{\alpha_t}
\end{equation}
In our greedy strategy we find the optimal blend parameter $w$ at each step between two trajectories of the PF-ODE,
the first trajectory being the model conditioned on $\bfz_a$ and the second being conditioned on $\bfz_b$.
This strategy is detailed in~\cref{eq:greedy-search} where $\mathcal{H}$ is the heuristic function which measures how ``optimal'' the morph is relative to the bona fide images. 
\begin{align}
    \label{eq:greedy-search}
    \boldsymbol\epsilon_t^{(w)} &= \mathrm{slerp}(\boldsymbol{\epsilon}_\theta(\bfx_t^{(ab)}, \bfz_a, t), \boldsymbol{\epsilon}_\theta(\bfx_t^{(ab)}, \bfz_b, t); w)\nonumber\\
    w_t' &= \argmin_{w \in \{w_n\}_{n=1}^B} \mathcal{H}\Big(\frac{1}{\alpha_t}\big(1 - \sigma_t \boldsymbol\epsilon_t^{(w)}\big), \bfx_0^{(a)}, \bfx_0^{(b)}\Big)
\end{align}
This results in a sequence of blend parameters $w_t'$ for each step in the PF-ODE solver.
Ultimately, the $\boldsymbol\epsilon_t^{(w_t')}$ for each timestep $t$ is used as the noise prediction when solving the PF-ODE.
We refer to this model as Greedy-DiM-S %
as it is the search variant of the Greedy-DiM family.

\begin{table}[h]
    \centering
    \caption{Comparison of search strategies with the identity loss as the heuristic function.}
    \footnotesize
    \begin{tabularx}{\linewidth}{lrrrr}
    \toprule
       && \multicolumn{3}{c}{\textbf{MMPMR}($\uparrow$)}\\
            \cmidrule(lr){3-5}
         \textbf{Search Strategy}                                 &   \textbf{NFE}($\downarrow$) & \textbf{AdaFace} &   \textbf{ArcFace} &  \textbf{ElasticFace}\\
    
    \midrule
     None                                                                       &350 &                 92.23 &               90.18 &                   93.05 \\
     Candidate                                                                                                 &2350& \textbf{95.91} &               92.84 &                   \textbf{95.5}  \\
     Greedy & 350 &                                                 95.71 &               \textbf{93.87} &                   95.3  \\
    \bottomrule
    \end{tabularx}
\label{tab:greedy-search}
\end{table}

In~\cref{tab:greedy-search} we compare the greedy search strategy to use no search strategy, as the case in the vanilla DiM approach in~\cite{blasingame_dim}, and to search across several morphed candidates generated by the diffusion process, as the strategy of Morph-PIPE~\cite{morph_pipe}.
Note both search strategies used $B = 21$ blends.
We observe that incorporating the information from the identity loss increases the effectiveness of the morphs.
Moreover, we notice a slight decrease in MMPMR when compared to searching the outputs of the diffusion process in two FR systems; however, there is an increase in MMPMR on the ArcFace FR system.
In comparison with searching across the morph candidates, this approach does not increase the NFE, as the greedy search is simply over different blends of the noise prediction at each timestep $t$, rather than the outputs of multiple diffusion processes.
\textit{N.B.}, the noise predictions of the two trajectories can be achieved in a single NFE by batching the inputs together.
See~\cref{appendix:nfe} for a detailed discussion of how we report NFE for DiM models in this paper.

\begin{table}[h]
    \centering
    \caption{Greedy search over $\{w_n\}_{n=1}^{21} \subseteq [0, 1]$ vs greedy gradient descent over $[0, 1]$.}
    \footnotesize   
    \begin{tabularx}{\linewidth}{@{\extracolsep{\fill}}lrrr}
    \toprule
       & \multicolumn{3}{c}{\textbf{MMPMR}($\uparrow$)}\\
            \cmidrule(lr){2-4}
         \textbf{Search Space}  & \textbf{AdaFace} &   \textbf{ArcFace} &  \textbf{ElasticFace}\\
    
    \midrule
     $\{w_n\}_{n=1}^{21} \subseteq [0, 1]$ & 95.71 &               93.87 &                   95.3  \\
    $[0,1]$ &               95.5  &               94.07 &                   95.09 \\
    \bottomrule
    \end{tabularx}
\label{tab:greedy-search-cont}
\end{table}

\noindent\textbf{Continuous Greedy Search.}
Next we replace the simple search over a fixed set of blend values $\{w_n\}_{n=1}^N$ to an optimization problem over a continuous set of blend values $[0, 1]$.
Obviously, it is intractable to use the same search strategy from earlier on this uncountably infinite set of blend values.
Rather we propose to use gradient descent to find the optimal $w \in [0, 1]$.
In~\cref{tab:greedy-search-cont} we provide the impact this choice has on MMPMR values. 
We observe that the continuous search space performs slightly worse than the discrete search space on the AdaFace and ElasticFace FR systems, but slightly stronger on the ArcFace FR system.
This mirrors the performance chance between the candidate and greedy search strategy shown in~\cref{tab:greedy-search}.

\begin{algorithm}[h]
    \caption{Greedy-DiM* Algorithm}
    \label{alg:greedy_dim}
    \small
    \begin{algorithmic}[1]
        \For {$n \gets N, N - 1, \ldots, 2$}
            \State $\boldsymbol{\epsilon} \gets \mathrm{stopgrad}(\boldsymbol\epsilon_\theta(\bfx_{t_n}^{(ab)}, \bfz_{ab}, t_n))$
            \State $\boldsymbol\epsilon^* \gets \boldsymbol\epsilon$
            \State $\mathcal{H}^* \gets \mathcal{H}(\hat\bfx_0, \bfx_0^{(a)}, \bfx_0^{(b)})$
            \For {$i \gets 1, 2, \ldots, n_{opt}$}
                \State $\hat\bfx_0 \gets \frac{1}{\alpha_{t_n}}(\bfx_{t_n}^{(ab)} - \sigma_{t_n}\boldsymbol\epsilon)$
                \State $\mathcal{H}(\boldsymbol\epsilon) \gets \mathcal{H}(\hat\bfx_0, \bfx_0^{(a)}, \bfx_0^{(b)})$
                \If {$\mathcal{H}(\boldsymbol\epsilon) < \mathcal{H}^*$}
                    \State $\mathcal{H}^* \gets \mathcal{H}(\boldsymbol\epsilon)$
                    \State $\boldsymbol\epsilon^* \gets \boldsymbol\epsilon$
                \EndIf
                \State $\boldsymbol{\epsilon} \gets \boldsymbol{\epsilon} - \eta \nabla_{\boldsymbol\epsilon} \mathcal{H}(\boldsymbol{\epsilon})$
            \EndFor
            \State $\bfx_{t_{n-1}}^{(ab)} \gets \Phi(\bfx_{t_n}^{(ab)}, \boldsymbol{\epsilon}^*, t_n)$ 
        \EndFor
        \State \textbf{return} $\bfx_0^{(ab)}$
    \end{algorithmic}
\end{algorithm}

\subsection{Greedy Optimization}
Rather than searching for the optimal blend $w$ of noise predictions that minimizes the heuristic function, we propose to directly optimize the noise prediction itself.
This is equivalent to optimize the step the ODE solver takes at each timestep $t$ towards some optimal $\bfx_0^*$ as defined by the heuristic function.
As we are now directly optimizing the predicted noise, it is no longer necessary to calculate the twin trajectories, instead the noise prediction network is only called a single time per timestep. 
In~\cref{alg:greedy_dim} we outline this greedy optimization strategy which we call Greedy-DiM* \textipa{["gri:di dIm sta\textturnr]}.
We provide a high-level overview of this algorithm in~\cref{fig:greedy_dim_overview}.
\cref{alg:greedy_dim} requires the original bona fide images $\bfx_0^{(a)}, \bfx_0^{(b)}$, the initial noise used in the generative process, $\bfx_T^{(ab)}$, the morphed conditional code $\bfz_{ab}$, the heuristic function $\mathcal{H}$ which measures how ``good'' the morphed images is, the number of optimizer steps, $n_{opt}$, the learning rate $\eta$, PF-ODE solver $\Phi$, and time schedule $\{t_n\}_{n=1}^N$.
We provide a PyTorch implementation of~\cref{alg:greedy_dim} with batch processing in~\cref{appendix:code}.

\begin{table}[h]
    \centering
    \caption{Comparison of Greedy-DiM-S and Greedy-DiM*}
    \footnotesize
    \begin{tabularx}{\linewidth}{lrrrr}
    \toprule
       && \multicolumn{3}{c}{\textbf{MMPMR}($\uparrow$)}\\
            \cmidrule(lr){3-5}
         \textbf{Algorithm}                                 &   \textbf{NFE}($\downarrow$) & \textbf{AdaFace} &   \textbf{ArcFace} &  \textbf{ElasticFace}\\
    
    \midrule
    Greedy-DiM-S & 350 & 95.71 &               93.87 &                   95.3  \\
    Greedy-DiM*                     &350&               \textbf{99.59} &              \textbf{ 99.18} &                   \textbf{99.39} \\
    \bottomrule
    \end{tabularx}
\label{tab:greedy-opt}
\end{table}

In~\cref{tab:greedy-opt} we compare Greedy-DiM-S to Greedy-DiM*. 
The greedy optimization strategy provides a \textit{significant} uplift in performance over the greedy search strategy while retaining the same number of NFE.
We posit that part of this is due to the enhanced size of the search space, allowing for a more optimal solution to be found.

\subsection{Theoretical Analysis}
Motivated by our strong empirical results, we present some theoretical justification  for the strong performance of Greedy-DiM* over Greedy-DiM-S and Morph-PIPE, along with some guarantees on the optimality of Greedy-DiM.
A greedy algorithm is one which chooses a locally optimal solution at each timestep; however, it is not automatically guaranteed that this locally optimal solution is globally optimal.
Famously, the greedy solution to the traveling salesman problem can actually yield the worst possible route~\cite{traveling_salesman}.
Motivated by this observation we present~\cref{thm:greed_is_good}.
\begin{theorem}
    \label{thm:greed_is_good}
    Given a sequence of monotonically descending timesteps, $\{t_n\}_{n=1}^N$, from $T$ to $0$, the DDIM solver to the Probability Flow ODE, and a heuristic function $\mathcal{H}$, the locally optimal solution admitted by Greedy-DiM* at time $t_n$ is globally optimal.
    \begin{proof}
        The proof is rather straightforward and based on the optimal substructure of optimizing the PF-ODE. We provide the full proof in~\cref{appendix:correctness_proof}.
    \end{proof}
\end{theorem}
\cref{thm:greed_is_good} shows that Greedy-DiM* does not just make locally optimal decisions, but globally optimal decisions.
Intuitively, as the noise prediction model can be restructured into an image prediction model, see~\cref{eq:x0_pred}, if the prediction is optimal at time $t$, then the same prediction at time $s < t$ is also optimal.
The result from~\cref{thm:greed_is_good} affirms that the choice to solve the optimization problem with a greedy strategy is principled and not merely rooted in experimental success.

\begin{figure}[h]
    \centering
    \includegraphics[trim={7cm 7cm 7cm 5cm},clip,width=\linewidth]{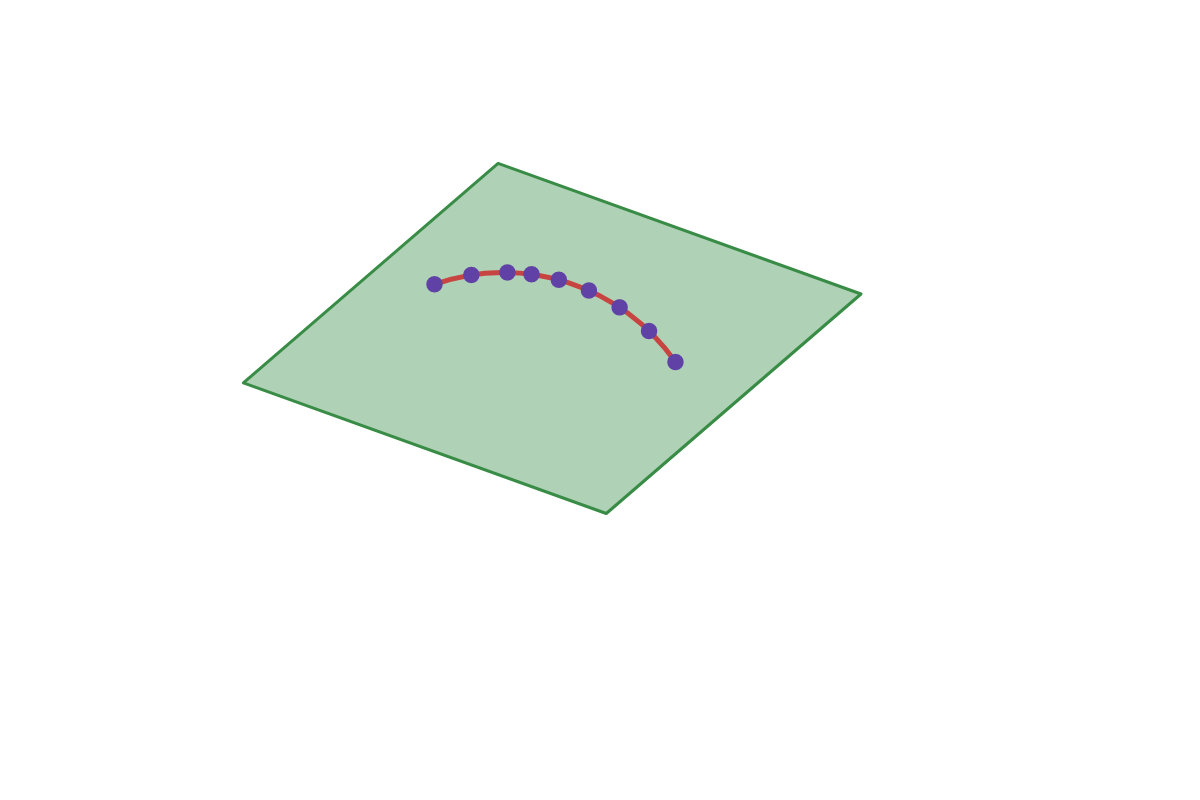}
    \caption{Illustration of the search space in $\R^2$ of different DiM algorithms at a single step. Purple denotes Morph-PIPE/Greedy-DiM-S, red denotes Greedy-DiM-S continuous, and green denotes Greedy-DiM*. Note the search spaces of the algorithms other than Greedy-DiM* lie in a low dimensional manifold.}
    \label{fig:search_space}
\end{figure}

\begin{figure*}[t]
    \centering
    \begin{subfigure}{0.14\textwidth}
        \includegraphics[width=0.98\textwidth]{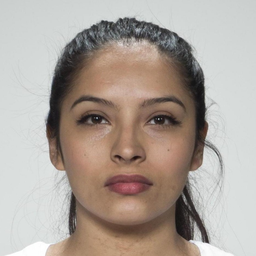}
    \end{subfigure}%
    \begin{subfigure}{0.14\textwidth}
        \includegraphics[width=0.98\textwidth]{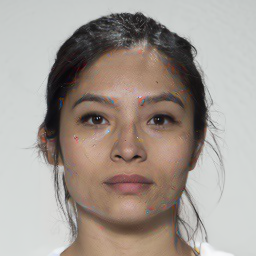}
    \end{subfigure}%
    \begin{subfigure}{0.14\textwidth}
        \includegraphics[width=0.98\textwidth]{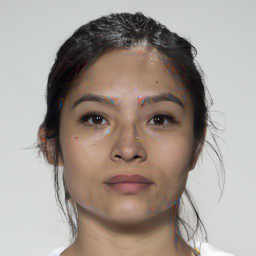}
    \end{subfigure}%
    \begin{subfigure}{0.14\textwidth}
        \includegraphics[width=0.98\textwidth]{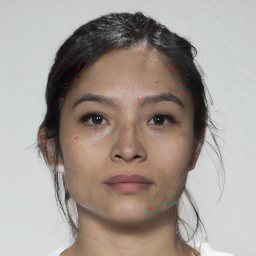}
    \end{subfigure}%
    \begin{subfigure}{0.14\textwidth}
        \includegraphics[width=0.98\textwidth]{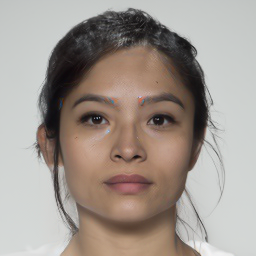}
    \end{subfigure}%
    \begin{subfigure}{0.14\textwidth}
        \includegraphics[width=0.98\textwidth]{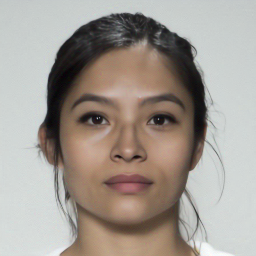}
    \end{subfigure}%
    \begin{subfigure}{0.14\textwidth}
        \includegraphics[width=0.98\textwidth]{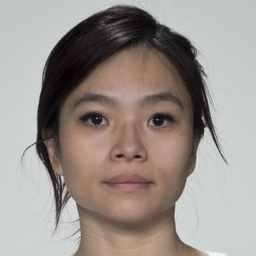}
    \end{subfigure}
    
    \begin{subfigure}{0.14\textwidth}
        \includegraphics[width=0.98\textwidth]{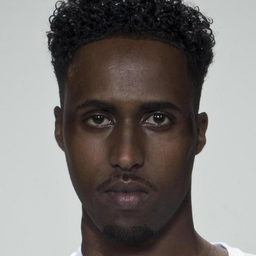}
        \caption{Identity $a$}
    \end{subfigure}%
    \begin{subfigure}{0.14\textwidth}
        \includegraphics[width=0.98\textwidth]{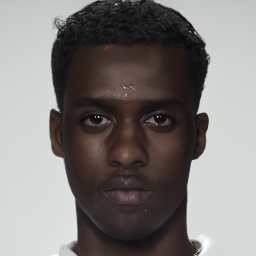}
        \caption{DiM-A}
    \end{subfigure}%
    \begin{subfigure}{0.14\textwidth}
        \includegraphics[width=0.98\textwidth]{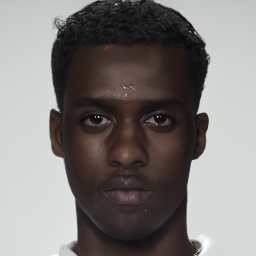}
        \caption{Fast-DiM}
    \end{subfigure}%
    \begin{subfigure}{0.14\textwidth}
        \includegraphics[width=0.98\textwidth]{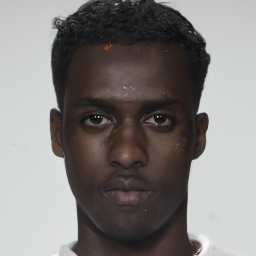}
        \caption{Morph-PIPE}
    \end{subfigure}%
    \begin{subfigure}{0.14\textwidth}
        \includegraphics[width=0.98\textwidth]{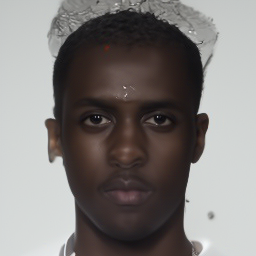}
        \caption{Greedy-DiM-S}
    \end{subfigure}%
    \begin{subfigure}{0.14\textwidth}
        \includegraphics[width=0.98\textwidth]{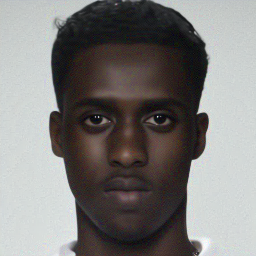}
        \caption{Greedy-DiM*}
    \end{subfigure}%
    \begin{subfigure}{0.14\textwidth}
        \includegraphics[width=0.98\textwidth]{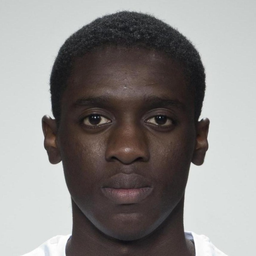}
        \caption{Identity $b$}
    \end{subfigure}
    \caption{Comparison of DiM morphs on the FRLL dataset.}
    \label{fig:morph_comp}
\end{figure*}

Next we explore the differences in the search spaces between Morph-PIPE, Greedy-DiM-S, and Greedy-DiM*.
As Morph-PIPE only explores a set of blend values on the inputs into the diffusion model, the search space is a discrete subset of $\X$ of size $B$.
The Greedy-DiM-S model has a vastly more expanded search space as there are $B$ different choices for each timestep, resulting in a finite search space with cardinality $B^N$.
Lastly, the search space of the Greedy-DiM* algorithm is $\X^N$.
We illustrate these differences in~\cref{fig:search_space}.
The search space of Greedy-DiM* is significantly larger than that of other algorithms as it takes full dimension rather than lying on a low dimensional manifold.
Recognizing the differences in the search spaces between the different algorithms, we consider the probability of the optimal solution being contained within the search space, from which we form~\cref{thm:greed_always_wins}.
\begin{theorem}
    \label{thm:greed_always_wins}
     Let $\pr$ be a probability distribution on a compact subset $\X \subseteq \R^n$ with full support on $\X$ which models the distribution of the optimal $\bfx_0^*$ and is absolutely continuous w.r.t. the $n$-dimensional Lebesgue measure $\lambda^n$ on $\X$.
     Let $\mathcal{S}_P, \mathcal{S}_S, \mathcal{S}^*$ denote the search spaces of the Morph-PIPE, Greedy-DiM-S, and Greedy-DiM* algorithms.
     Then the following statements are true.
     \begin{enumerate}
         \item $\pr(\mathcal{S}_P) = \pr(\mathcal{S}_S) = 0$.
         \item $\pr(\mathcal{S}^*) = 1$.
     \end{enumerate}
    \begin{proof}
        The proof is based on classic results from measure theory. The full proof is provided in~\cref{appendix:proof_solution_exists}.
    \end{proof} 
\end{theorem}
\cref{thm:greed_always_wins} shows that the search space of all algorithms other than Greedy-DiM* \textit{almost surely} do not contain the optimal solution.
The results from~\cref{thm:greed_always_wins} further justifies the superior performance exhibited by Greedy-DiM* in relation to other search strategies.
Whereas Morph-PIPE and Greedy-DiM-S search over a discrete subset of a fixed trajectory between the two bona fide images, the Greedy-DiM* algorithm searches over the whole image space at each iteration, allowing Greedy-DiM* to find an optimal solution that deviates from the fixed trajectory.

\section{Experimental Setup}
\label{sec:experiemental_setup}

\subsection{Datasets}
The open-source SYN-MAD 2022 IJCB competition dataset~\cite{syn-mad22} is chosen as the dataset to benchmark the proposed morphing attacks on, 
as it has been used as a common benchmark for measuring the performance of face morphing attacks and MAD algorithms~\cite{fast_dim,justify_synmad_1,justify_synmad_2}.
The SYN-MAD 2022 dataset is derived from the Face Research Lab London (FRLL) dataset~\cite{frll}.
FRLL is a dataset of high-quality captures of 102 different individuals with frontal images and neutral lighting.
There are two images per subject, an image of a ``neutral'' expression and one of a ``smiling'' expression.
The ElasticFace~\cite{elasticface} FR system was used to select the top 250 most similar pairs, in terms of cosine similarity, of bona fide images for both genders, resulting in a total of 500 bona fide image pairs for face morphing~\cite{syn-mad22}.
The SYN-MAD 2022 dataset includes morphed images with three landmark-based attacks along with MIPGAN-I and MIPGAN-II attacks.
The three landmark-based attacks are the open-source OpenCV attack, the commercial-of-the-shelf (COTS) FaceMorpher, and online-tool Webmorph~\cite{syn-mad22}.
\textit{N.B.,} the FaceMorpher attack in the SYN-MAD 2022 is not the same attack used in~\cite{blasingame_dim,sebastien_gan_threaten} which was a different open-source landmark-based attack of the same name.
The SYN-MAD 2022 dataset includes the OpenCV, FaceMorpher, Webmorph, MIPGAN-I, and MIPGAN-II attacks.
In addition to the five attacks provided by the SYN-MAD 2022, we used the same 500 pairs to generate five DiM attacks for comparison purpose. This includes the DiM algorithm using variants A and C from~\cite{blasingame_dim}, the Morph-PIPE algorithm from~\cite{morph_pipe}, and the Fast-DiM and Fast-DiM-ode algorithms from concurrent work~\cite{fast_dim}.
Further details can be found in~\cref{appendix:datasets}.

\subsection{FR Systems}
We evaluate the morphing attacks against three publicly available FR systems, namely the ArcFace~\cite{deng2019arcface}, AdaFace~\cite{adaface}, and ElasticFace~\cite{elasticface} models.
We chose to evaluate these three models as they represent a combination of widely used and state-of-the-art FR systems which have been used by other works to study the effectiveness of morphing attacks~\cite{blasingame_dim,morph_pipe,fast_dim}.
\textit{N.B.}, the ArcFace model used in the identity loss is not the same FR system used during evaluation.
Further details on the configuration of the FR systems are found in~\cref{appendix:fr_config}.

\subsection{Metrics}
To measure the effectiveness of our proposed morphing attack, we measure the Mated Morph Presentation Match Rate (MMPMR) metric proposed by Scherhag~\etal~\cite{mmpmr}.
The MMPMR measures the number of morphed images that successfully trick an FR system into falsely matching it with each of the contributing subjects at a given verification threshold.
The metric is defined as
\begin{equation}
    M(\delta) = \frac{1}{M} \sum_{n=1}^M \bigg\{\bigg[\min_{n \in \{1,\ldots,N_m} S_m^n\bigg] > \delta\bigg\}
\end{equation}
where $\delta$ is the verification threshold, $S_m^n$ is the similarity score of the $n$-th subject of morph $m$, $N_m$ is the total number of contributing subjects to morph $m$, and $M$ is the total number of morphed images.
In our reporting of the MMPMR metric we set the verification threshold such that the False Match Rate (FMR) is $0.1\%$.

We also report the Morphing Attack Potential (MAP), an extension on the MMPMR metric proposed by Ferrara~\etal~\cite{map_metric}.
The MAP metric aims to provide a more comprehensive assessment of the risk a morphing attack poses to FR systems.
This is accomplished by reporting a matrix such that $\mathrm{MAP}[r, c]$ denotes the proportion of morphed images that successfully register a false accept against at least $r$ attempts against each contributing subject of at least $c$ FR systems.
As the SYN-MAD 2022 dataset only has a single probe image per subject, for the other image was used in the creation of the face morph and thus not suitable to be a probe image during evaluation, we simply report $\mathrm{MAP}[1, c]$.
This can still provide useful insight into the generality of a morphing attack.

\section{Results}
In this section we compare our proposed Greedy-DiM algorithms to existing DiM and concurrent algorithms in addition to several GAN and landmark-based morphing attacks, studying a total of twelve morphing attacks of which ten are from other works.
The specific hyperparameter configurations used in these experiments for the Greedy-DiM-S and Greedy-DiM* algorithms are noted in~\cref{appendix:hyperparams}.
\cref{fig:morph_comp} provides some examples of morphed images generated by different DiM algorithms.
We observe that all DiM algorithms other than Greedy-DiM* exhibit strange saturation artefacts, \ie, the generally bright red splotches on the morphed images, a result of clipping the dynamic range of the image to [-1, 1].
The Greedy-DiM-S algorithm exhibit some blend artefacting outside the core face region unlike the other algorithms.

\begin{table}[t]
\caption{Vulnerability of different FR systems across different morphing attacks on the SYN-MAD 2022 dataset. FMR = 0.1\%.}
\centering
\footnotesize
\begin{tabularx}{\linewidth}{@{\extracolsep{\fill}}lrrrr}
\toprule
&& \multicolumn{3}{c}{\textbf{MMPMR}($\uparrow$)}\\
            \cmidrule(lr){3-5}
 \textbf{Morphing Attack}                                 & \textbf{NFE($\downarrow$)}  &\textbf{AdaFace} &   \textbf{ArcFace} &  \textbf{ElasticFace} \\
\midrule
 FaceMorpher~\cite{syn-mad22}                                     &-&               89.78 &               87.73 &                   89.57 \\
 Webmorph~\cite{syn-mad22}                                        &-&               97.96 &               96.93 &                   98.36 \\
 OpenCV~\cite{syn-mad22}                                          &-&               94.48 &               92.43 &                   94.27 \\
 MIPGAN-I~\cite{mipgan}                                      &-&               72.19 &               77.51 &                   66.46 \\
 MIPGAN-II~\cite{mipgan}                                       &-&               70.55 &               72.19 &                   65.24 \\
 DiM-A~\cite{blasingame_dim}                                           &350&               92.23 &               90.18 &                   93.05 \\
 DiM-C~\cite{blasingame_dim}                                           &350&               89.57 &               83.23 &                   86.3  \\
Fast-DiM~\cite{fast_dim} &300&               92.02 &               90.18 &                   93.05 \\ 
Fast-DiM-ode~\cite{fast_dim}   &150&    91.82 &               88.75 &                   91.21    \\     
 Morph-PIPE~\cite{morph_pipe} & 2350                                                                             &               95.91 &               92.84 &                   95.5  \\
\textbf{Greedy-DiM-S} & 350 &                                                 95.71 &               93.87 &                   95.3  \\
\textbf{Greedy-DiM*}      & 270 &               \textbf{100}  &              \textbf{100}    &                   \textbf{100}  \\
\bottomrule
\end{tabularx}
\label{tab:mmpmr}
\end{table}

\subsection{Vulnerability Study}
In~\cref{tab:mmpmr} we measure the MMPMR of the twelve studied morphing attacks on all three FR systems in addition to reporting the NFE if applicable.
Our results show that Greedy-DiM* is unreasonably effective, achieving a 100\% MMPMR across the board, boasting superior performance to all other morphing attacks.
We observe that Greedy-DiM* is the \textit{first} representation-based morphing attack to \textit{consistently} outperform landmark-based morphing attacks~\cite{blasingame_dim,morph_pipe,fast_dim,mipgan,syn-mad22}.
Remarkably, this performance is achieved with a reduction in NFE, having the lowest NFE out of all proposed DiM models with the exception of Fast-DiM-ode.
As expected, Greedy-DiM-S under-performs Greedy-DiM*, maintaining similar performance to Morph-PIPE, but with significantly lower NFE.
Lastly, mirroring the observations in~\cite{fast_dim,syn-mad22} we notice the MIPGAN algorithms tend to perform very poorly and that Webmorph performs quite well.

\begin{table}[h]
    \centering
    \caption{MAP$(\uparrow)$ metric for all three FR systems on the SYN-MAD 2022 dataset. FMR = 0.1\%.}
    \footnotesize
    \begin{tabularx}{\linewidth}{@{\extracolsep{\fill}}lrrrr}
    \toprule
    && \multicolumn{3}{c}{\textbf{Number of FR Systems}}\\
    \cmidrule(lr){3-5}
    \textbf{Morphing Attack} & \textbf{NFE($\downarrow$)} & \textbf{1} & \textbf{2} & \textbf{3}\\
    \midrule
     FaceMorpher~\cite{syn-mad22}                                     &- & 92.23 & 89.57 & 85.28 \\
     Webmorph~\cite{syn-mad22}                                        &-& 98.77 & 98.36 & 96.11 \\
     OpenCV~\cite{syn-mad22}                                          &-& 97.55 & 93.87 & 89.78 \\
     MIPGAN-I~\cite{mipgan}                                        &-& 85.07 & 72.39 & 58.69 \\
     MIPGAN-II~\cite{mipgan}                                       &-& 80.37 & 69.73 & 57.87 \\
     DiM-A~\cite{blasingame_dim}                                           &350& 96.93 & 92.43 & 86.09 \\
     DiM-C~\cite{blasingame_dim}                                           &350& 92.84 & 87.53 & 78.73 \\
     Fast-DiM~\cite{fast_dim}   &300& 97.14 & 92.43 & 85.69 \\
     Fast-DiM-ode~\cite{fast_dim}   &150& 95.91 & 91.21 & 84.66 \\
     Morph-PIPE~\cite{morph_pipe} & 2350 &  98.16 &  95.71 &  90.39\\
     \textbf{Greedy-DiM-S} & 350 &  97.34 &  95.71 &  91.82 \\
     \textbf{Greedy-DiM*} & 270 & \textbf{100}    & \textbf{100}    & \textbf{100}    \\
    \bottomrule
    \end{tabularx}
    \label{tab:map}
\end{table}

Similar conclusions are drawn from~\cref{tab:map} which illustrates the $\mathrm{MAP}[1, c]$ metric for all twelve studied morphing attacks.
Again, Greedy-DiM* performs unreasonably well having a 100\% chance to fool all three FR systems studied in this work.
The next closest attack, is Webmorph with a 96.11\% chance to fool all three FR systems, and the nearest representation-based attack is Greedy-DiM-S with a 91.82\% chance to fool all three FR systems.
Likewise, the MIPGAN attacks perform abysmally, with only a mere $<60\%$ chance to fool all three FR systems.

\begin{table*}[t]
    \centering
    \caption{Detection Study on all training subsets of the SYN-MAD 2022 dataset}
    \footnotesize
    \begin{tabularx}{\linewidth}{@{\extracolsep{\fill}}lrrrrrrrrrrrr}
    \toprule
     &\multicolumn{4}{c}{\textbf{OpenCV}}
     &\multicolumn{4}{c}{\textbf{MIPGAN-II}}
     &\multicolumn{4}{c}{\textbf{DiM-A}}\\
     \cmidrule(lr){2-5}
     \cmidrule(lr){6-9}
     \cmidrule(lr){10-13}
     &&\multicolumn{3}{c}{\textbf{MACER @ BPCER($\uparrow$)}}
     &&\multicolumn{3}{c}{\textbf{MACER @ BPCER($\uparrow$)}}
     &&\multicolumn{3}{c}{\textbf{MACER @ BPCER($\uparrow$)}}\\
     \cmidrule(lr){3-5}
     \cmidrule(lr){7-9}
     \cmidrule(lr){11-13}
     \textbf{Morphing Attack} &
     \textbf{EER($\uparrow$)} & \textbf{0.1\%} & \textbf{1.0\%} & \textbf{5.0\%} &
     \textbf{EER($\uparrow$)} & \textbf{0.1\%} & \textbf{1.0\%} & \textbf{5.0\%} &
     \textbf{EER($\uparrow$)} & \textbf{0.1\%} & \textbf{1.0\%} & \textbf{5.0\%}\\
    \midrule

 FaceMorpher\cite{syn-mad22}                                                                                             & 31.86 &                99.02 &                97.06 &                80.39 & \textbf{82.84} &               \textbf{100}    &               \textbf{100}    &                \textbf{99.51} & 44.61 &                99.02 &                97.06 &                87.75 \\
 Webmorph~\cite{syn-mad22}                                                                                                &  0.49 &                 0.98 &                 0.49 &                 0.49 & 51.47 &               \textbf{100}    &                99.51 &                93.14 & 36.27 &                98.04 &                90.69 &                74.02 \\
 OpenCV~\cite{syn-mad22}                                                                                                  &  0.98 &                 2.45 &                 0.49 &                 0.49 & 50.49 &                99.51 &                95.1  &                90.69 & 39.71 &                99.02 &                85.29 &                73.04 \\
 MIPGAN-I~\cite{mipgan}                                                                                                & 24.02 &                83.82 &                70.1  &                49.02 &  8.33 &                38.24 &                24.51 &                14.22 & 44.61 &                \textbf{99.51} &                95.1  &                85.78 \\
 MIPGAN-II~\cite{mipgan}                                                                                               & 26.47 &                73.53 &                59.8  &                45.59 &  3.92 &                34.31 &                 8.82 &                 2.94 & \textbf{52.94} &                99.02 &                \textbf{98.04} &                \textbf{90.69} \\
  DiM-A~\cite{blasingame_dim}                                                                                                   & \textbf{82.84} &               \textbf{100}    &               \textbf{100}    &               \textbf{100}    & 54.9  &                99.51 &                99.51 &                93.63 &  0.98 &                 2.94 &                 0.98 &                 0    \\
 DiM-C~\cite{blasingame_dim}                                                                                                   & 63.73 &                99.51 &                99.02 &                98.53 & 43.63 &                99.51 &                93.63 &                85.29 &  5.39 &                42.65 &                19.61 &                 6.37 \\
Fast-DiM~\cite{fast_dim}                                                         & 75    &               \textbf{100}    &               \textbf{100}    &               \textbf{100}    & 50    &                99.51 &                99.51 &                94.61 &  2.94 &                12.25 &                 5.88 &                 1.96 \\
Fast-DiM-ode~\cite{fast_dim}                                                           & 75.49 &               \textbf{100}    &               \textbf{100}    &               \textbf{100}    & 56.37 &                99.51 &                99.51 &                98.53 &  1.96 &                 7.35 &                 2.94 &                 0.49 \\
 Morph-PIPE~\cite{morph_pipe}                                                                        & 82.35 &               \textbf{100}    &               \textbf{100}    &               \textbf{100}    & 54.9  &                99.51 &                99.51 &                93.63 &  0.98 &                 5.39 &                 0.98 &                 0    \\
 \textbf{Greedy-DiM-S}                                            & 52.94 &               \textbf{100}    &               \textbf{100}    &                98.53 & 36.76 &               \textbf{100}    &                92.65 &                79.41 &  6.37 &                63.73 &                14.22 &                 6.86 \\
\textbf{Greedy-DiM*} & 37.75 &                99.02 &                93.63 &                85.29 & 34.31 &                96.57 &                93.14 &                75    & 17.16 &                85.78 &                56.37 &                42.65 \\
    \bottomrule
    \end{tabularx}
    \label{tab:det_study}
\end{table*}
\subsection{Detectability Study}
\label{sec:detect_study}
In this section we study the detectability of Greedy-DiM by implementing a Single-image MAD (S-MAD) detector trained on various morphing attacks.
The detectability study follows a similar approach to that of~\cite{blasingame_dim,fast_dim}.
We employ stratified $k$-fold cross validation when performing the study to ensure a consistent balance of morphed and bona fide images in each fold. 
Further details on the S-MAD detector and other configuration details for this study can be found in~\cref{appendix:detect}.
The S-MAD model is trained in three different scenarios for a potential S-MAD algorithm.
In the first scenario we train the S-MAD detector on OpenCV morphs, the second MIPGAN-II morphs, and the third DiM-A morphs.

In order to assess the detectability of morphing attacks by the S-MAD algorithm, we measure the Equal Error Rate (EER), Morphing Attack Presentation Error Rate (MACER), and Bona fide Presentation Classification Error Rate (BPCER).
The results of our detectability are shown in~\cref{tab:det_study}.
In general, we notice that in order for the S-MAD algorithm to be able to \textit{consistently} detect morphing attacks it needs to be trained on that general type of morphing attack, \ie, trained on an attack which uses a similar generation process in creating the morphed images.
When the S-MAD detector is trained on OpenCV morphs the Greedy-DiM attacks are the most likely to be detected out of all the DiM attacks; however, the MIPGAN and other landmark-based attacks are more easily detected.
We notice that the detector trained on MIPGAN-II has a hard time detecting the attack it was trained on, much less detecting the other non-MIPGAN attacks.
Interestingly, we observe that for the S-MAD detector trained on DiM-A that the Greedy-DiM attacks are the hardest amongst the DiM attacks to detect with Morph-PIPE being the easiest, outside of the trivial case of DiM-A.
This highlights the substantial difference between Greedy-DiM, and Greedy-DiM* in particular, and all other DiM attacks.

\section{Conclusions}
We propose Greedy-DiM, a novel family of face morphing algorithm that use greedy strategies to enhance the optimization of the PF-ODE solver.
Greedy-DiM* achieves unreasonably effective performance resulting in a 100\% MMPMR on all tested FR systems, far surpassing previous efforts.
Moreover, this is accomplished with minimal overhead.
Additional theoretical analysis corroborates the strong experimental results.
While this work has focused primarily on face morphing, the proposed greedy strategies can be used for a variety of heuristic functions and tasks, enabling heuristic guided diffusion for other applications.

In this work we only evaluated our morphs on the SYN-MAD 2022 dataset and used only a single diffusion backbone 
rather than exploring multiple diffusion architectures.
While not tested in this work, it should be a rather straightforward extension to apply these techniques to Latent Diffusion Models.
Moreover, modifications to the heuristic function could be explored such as encouraging the morphed image to be less ``detectable''.
Additionally, the scheduling of the greedy procedure was left as a linear schedule.
Such extensions are left for future work.

\noindent\textbf{Reproducibility Statement.} To ensure the reproducibility and completeness of this work we include the Appendix with five main sections.
In~\cref{appendix:proofs} we provide the complete proofs for all theorems in this paper.
In~\cref{appendix:implement} we provide the implementation details for the proposed algorithms.
Likewise,~\cref{appendix:eval_details} provides further detail on how we performed our evaluations of the proposed algorithms.
In~\cref{appendix:addtional_results} we provide additional results that were outside the scope of the main paper that the reader may find interesting.
Lastly, in~\cref{appendix:additional_images} we provide additional samples created by the proposed algorithms.

\noindent\textbf{Ethics Statement.}
Face morphs can be used for a variety of malicious purposes from creating misleading content to attacking FR systems.
Our advances in heuristic guided generation of diffusion models can be harnessed to produce deceptive face morphs or ``deepfakes'' among other attacks.
We hope our colleagues will incorporate this work into future research on MAD algorithms to mitigate the large threat these morphs pose to unprotected FR systems.

\section*{Acknowledgements}
This material is based upon work supported by the Center for Identification Technology Research and National Science Foundation under Grant \#1650503.

{\small
\bibliographystyle{IEEEtran}
\bibliography{bib}
}

\clearpage
\onecolumn

\tableofcontents
\clearpage

\appendix
\section{Proofs}
\label{appendix:proofs}

\subsection{Greedy-DiM is Globally Optimal}
\label{appendix:correctness_proof}
\begin{theorem}
    Given a sequence of monotonically descending timesteps, $\{t_n\}_{n=1}^N$, from $T$ to $0$ and the DDIM solver to the Probability Flow ODE in the Variance-Preserving formulation and a heuristic function $\mathcal{H}$, the locally optimal solution $\boldsymbol\epsilon_{t_n}$ which minimizes $\mathcal{H}$ at timestep $t_n$ is globally optimal.
    \begin{proof}
        Let $\bfx_0'$ minimize $\mathcal{H}$. Consider an arbitrary timestep $t$ with a locally optimal $\boldsymbol\epsilon_t'$ which is related to $\bfx_0'$ by
        \begin{equation}
            \label{eq:eps_t_proof}
            \boldsymbol\epsilon_t' = \frac{\bfx_t - \alpha_t \bfx_0'}{\sigma_t}
        \end{equation}
        \ie, $\boldsymbol\epsilon_t$ minimizes $\mathcal{H}$ at time $t$.
        We will show that for any $s < t$, $\boldsymbol\epsilon_t = \boldsymbol\epsilon_s$.
        Let $s$ be the next timestep in the sequence.
        The DDIM update equation shows that $\bfx_s$ is defined as
        \begin{equation}
            \label{eq:update_ddim_proof}
            \bfx_s = \frac{\alpha_s}{\alpha_t}(\bfx_t - \sigma_t \boldsymbol\epsilon_t') + \sigma_s \boldsymbol\epsilon_t'
        \end{equation}
        Plugging~\cref{eq:eps_t_proof} into~\cref{eq:update_ddim_proof} we have
        \begin{equation}
            \label{eq:update_ddim_proof_2}
            \bfx_s = \alpha_s\bfx_0' + \frac{\sigma_s}{\sigma_t}(\bfx_t - \alpha_t \bfx_0')
        \end{equation}
        Now we write~\cref{eq:eps_t_proof} at time $s$ in terms of~\cref{eq:update_ddim_proof_2} which yields
        \begin{equation}
            \label{eq:eps_s_proof}
            \boldsymbol\epsilon_s' = \frac{\alpha_s\bfx_0' + \frac{\sigma_s}{\sigma_t}(\bfx_t - \alpha_t \bfx_0') - \alpha_s\bfx_0'}{\sigma_s}
        \end{equation}
        Simplifying,~\cref{eq:eps_s_proof} we find
        \begin{equation}
            \boldsymbol\epsilon_s' = \frac{\bfx_t - \alpha_t\bfx_0'}{\sigma_t} = \boldsymbol\epsilon_t'
        \end{equation}
        Therefore for any timestep $s < t$ the locally optimal solution $\boldsymbol\epsilon_t'$ is locally optimal at time $s$.
        Thus starting at time $T$ with locally optimal solution $\boldsymbol\epsilon_T'$ the locally optimal solutions for all $t \in [0, T)$ are $\boldsymbol\epsilon_T'$.
        Therefore, for any timestep $t_n$ the locally optimal solution $\boldsymbol\epsilon_{t_n}$ is globally optimal.
    \end{proof}
\end{theorem}

\subsection{Probability of Finding the Optimal Solution}
\label{appendix:proof_solution_exists}

\begin{theorem}
     Let $\pr$ be a probability distribution on a compact subset $\X \subseteq \R^n$ with full support on $\X$ which models the distribution of the optimal $\bfx_0^*$ and is absolutely continuous w.r.t. the $n$-dimensional Lebesgue measure $\lambda^n$ on $\X$.
     Let $\mathcal{S}_P, \mathcal{S}_S, \mathcal{S}^*$ denote the search spaces of the Morph-PIPE, Greedy-DiM-S, and Greedy-DiM* algorithms.
     Then the following statements are true.
     \begin{enumerate}
         \item $\pr(\mathcal{S}_P) = \pr(\mathcal{S}_S) = 0$.
         \label{proof:stmnt_bad_search}
         \item $\pr(\mathcal{S}^*) = 1$.
         \label{proof:stmnt_good_search}
     \end{enumerate}
    \begin{proof}
        Let $\bfx_T^{(a)}, \bfx_T^{(b)}$ denote the initial noise of the bona fide images and let $\bfz_a, \bfz_b$ denote the conditional representations of the bona fide images.
        Remark a measure $\mu$ is said to be absolutely continuous w.r.t. to $\nu$ if and only if for all measurable sets $A$, $\nu(A) = 0 \implies \mu(A) = 0$
        We consider the search space of the Morph-PIPE algorithm $\mathcal{S}_P$. We can construct $\mathcal{S}_P$ as
        \begin{equation}
            \mathcal{S}_P = \bigg\{ \Phi^N(\mathrm{slerp}(\bfx_T^{(a)}, \bfx_T^{(b)}; w), \mathrm{lerp}(\bfz_a, \bfz_b; w)) \;\Big |\; w \in \{w_n\}_{n=1}^B \bigg\}
        \end{equation}
        where $\Phi^N(\bfx_T, \bfz)$ denotes the output of the diffusion model with $N$ sampling steps and $\{w_n\}_{n=1}^B \subseteq [0,1]$ is set of blend values.
        By construction it follows that $|\mathcal{S}_P| = B$, \ie, the $B$ candidates generated by the Morph-PIPE algorithm.
        Clearly, search space is a null set w.r.t. to the Lebesgue measure on $\X$ and therefore $\lambda^n(\mathcal{S}_P) = 0$ which by the definition of absolutely continuity implies $\pr(\mathcal{S}_P) = 0$.

        Next we consider the search space of the Greedy-DiM-S algorithm $\mathcal{S}_S$.
        Clearly the search space at a timestep $t$, is a set of a size of the $B$ blend values explored.
        Over the $N$ sampling steps of the diffusion model the greedy algorithm can explore a maximum of $B^N$ models.
        It follows the total search space $\mathcal{S}_S$ is still discrete and thus the Lebesgue measure of the search space is zero which implies that $\pr(\mathcal{S}_S) = 0$.
        We have now shown that Statement~\ref{proof:stmnt_bad_search} is true.

        Lastly, we consider the search space of the Greedy-DiM* algorithm $\mathcal{S}^*$. 
        By definition Greedy-DiM* performs gradient descent on $\X$ at each timestep and therefore the search space of the whole algorithm is $\mathcal{S}^* = \X$.
        By definition $\pr(\X) = \pr(\mathcal{S}^*) = 1$ as $\pr$ takes full support on $\X$.
        We have now shown that Statement~\ref{proof:stmnt_good_search} is true finishing the proof.
    \end{proof}
\end{theorem}

Note it can be shown that search space of Greedy-DiM-S continuous lies on a low-dimensional manifold of $\X$ and therefore takes probability $0$, but the proof is more technical so we opt to omit it as it is not important to our analysis.

\section{Implementation Details}
\label{appendix:implement}

\subsection{DiM Algorithm}
\label{appendix:dim_alg}
For completeness we provide the DiM algorithm from~\cite{blasingame_dim} in our own notation in~\cref{alg:dim_morph}.
The original bona fide images are denoted $\bfx_0^{(a)}$ and $\bfx_0^{(b)}$.
The conditional encoder is $\mathcal{E}: \X \to \Z$, $\Phi$ is the numerical PF-ODE solver, $\Phi^+$ is the numerical ODE solver of the PF-ODE as time runs forwards from $0$ to $T$.

\begin{algorithm}[h]
    \caption{DiM Framework.}
    \label{alg:dim_morph}
    \begin{algorithmic}[1]
        \Require{Blend parameter $w = 0.5$. Time schedule $\{t_i\}_{i=1}^N \subseteq [0, T], t_i < t_{i+1}$.}
        \State $\bfz_a \gets \mathcal{E}(\bfx_0^{(a)})$ \Comment{Encoding bona fides into conditionals.}
        \State $\bfz_b \gets \mathcal{E}(\bfx_0^{(b)})$
        \For {$i \gets 1, 2, \ldots, N - 1$}
            \State $\bfx_{t_{i+1}}^{(a)} \gets \Phi^{+}(\bfx_{t_i}^{(a)}, \boldsymbol{\epsilon}_\theta(\bfx_{t_i}^{(a)}, \bfz_a, t_i), t_i)$ \Comment{Solving the PF-ODE as time runs from $0$ to $T$.}
            \State $\bfx_{t_{i+1}}^{(b)} \gets \Phi^{+}(\bfx_{t_i}^{(b)}, \boldsymbol{\epsilon}_\theta(\bfx_{t_i}^{(b)}, \bfz_b, t_i), t_i)$ 
        \EndFor
        \State $\bfx_T^{(ab)} \gets \textrm{slerp}(\bfx_T^{(a)}, \bfx_T^{(b)}; w)$ \Comment{Morph initial noise.}
        \State $\bfz_{ab} \gets \textrm{lerp}(\bfz_a, \bfz_b; w)$ \Comment{Morph conditionals.}
        \For {$i \gets N, N - 1, \ldots, 2$}
            \State $\bfx_{t_{i-1}}^{(ab)} \gets \Phi(\bfx_{t_i}^{(ab)}, \boldsymbol{\epsilon}_\theta(\bfx_{t_i}^{(ab)}, \bfz_{ab}, t_i), t_i)$ \Comment{Solving the PF-ODE as time runs from $T$ to $0$.}
        \EndFor
        \State \textbf{return} $\bfx_0^{(ab)}$
    \end{algorithmic}
\end{algorithm}

Note for a vector space $V$ and two vectors $u, v \in V$, the spherical interpolation by a factor of $\gamma$ is given as
\begin{equation}
    \textrm{slerp}(u, v; \gamma) = \frac{\sin((1 - \gamma) \theta)}{\sin \theta}u + \frac{\sin(\gamma \theta)}{\sin \theta}v
\end{equation}
where
\begin{equation}
\theta = \frac{\arccos(u \cdot v)}{\|u\| \, \|v\|}
\end{equation}

\subsection{Repositories Used}
\label{appendix:repos}
For reproducibility purposes we provide a list of links to the official repositories of other works used in this paper.
\begin{enumerate}
    \item The SYN-MAD 2022 dataset used in this paper can be found at \url{https://github.com/marcohuber/SYN-MAD-2022}.
    \item The implementation for OpenCV morphs can be found at \url{https://learnopencv.com/face-morph-using-opencv-cpp-python}.
    \item The COTS FaceMorpher tool can be found at \url{https://www.luxand.com/facemorpher}.
    \item The Webmorph online tool can be found at \url{https://webmorph.org}.
    \item The ArcFace models, MS1M-RetinaFace dataset, and MS1M-ArcFace dataset can be found at \url{https://github.com/deepinsight/insightface}.
    \item The ElasticFace model can be found at \url{https://github.com/fdbtrs/ElasticFace}.
    \item The AdaFace model can be found at \url{https://github.com/mk-minchul/AdaFace}.
    \item The official Diffusion Autoencoders repository can be found at \url{https://github.com/phizaz/diffae}.
    \item The official MIPGAN repository can be found at \url{https://github.com/ZHYYYYYYYYYYYY/MIPGAN-face-morphing-algorithm}.
    \item The SE-ResNeXt101-32x4d can be found at \url{https://catalog.ngc.nvidia.com/orgs/nvidia/resources/se_resnext_for_pytorch}.
\end{enumerate}

\subsection{Hyperparameters}
\label{appendix:hyperparams}
In~\cref{tab:params} we enumerate the hyperparameters used in the main paper for the Greedy-DiM algorithms and provide some justification for our choices.
For Greedy-DiM-S we used the recommended $N = 100$ steps for the DDIM solver from~\cite{blasingame_dim,diffae}.
In our experiments we found $N = 20$ to perform just as well as $N = 100$ for Greedy-DiM*, see~\cref{tab:n_steps}.
We use the recommend $N_F = 250$ steps with the DiffAE forward ODE solver~\cite{blasingame_dim,diffae}.
For the Greedy-DiM-S algorithm we use $B = 21$ as that is what was used in the Morph-PIPE piper~\cite{morph_pipe}.
Additionally, we use slerp as our interpolation function for the predicted noise which is a common choice when morphing in the image domain~\cite{blasingame_dim,diffae}.
At the suggestion of Song~\etal~\cite{song2023consistency} we use the Rectified Adam optimizer~\cite{liu2019radam}, with no learning rate decay or warm-up.
In we performed an initial parameter study and found a learning rate of $0.01$ to work well and chose to use that for the remainder of our experiments.
We performed a small parameter search on the momentum parameters of the optimizer and found the values $\beta_0 = 0.5$ and $\beta_1 = 0.9$ recommended by Gu~\etal~\cite{weird_betas} to work quite well, see~\cref{tab:betas}.
We varied the optimizer stride, \ie, how many steps to skip the greedy search on, and found it to negatively impact results so we opted to leave it 1, thereby disabling optimizer striding.
In our initial experiments we found $n_{opt} = 50$ to work quite well.
We performed an extensive study on the heuristic function and found the ArcFace identity loss to work the best. See~\cref{tab:heuristic} for the full results of this study.

\begin{table}[h]
    \centering
    \caption{Hyperparamters used for the main paper.}
    \begin{tabularx}{\linewidth}{@{\extracolsep{\fill}}lll}
    \toprule
    \textbf{Hyperparameter} & \textbf{Greedy-DiM-S} & \textbf{Greedy-DiM*}\\
    \midrule
    \textbf{ODE Solvers}\\
    $N$ & 100 & 20\\
    ODE Solver & DDIM & DDIM \\
    $N_F$ & 250 & 250\\
    Forward ODE Solver & DiffAE & DiffAE\\
    \midrule
    \textbf{Search}\\
    Number of blends & 21 & -\\
    Interpolation function & slerp & -\\
    \midrule
    \textbf{Optimization}\\
    Optimizer & - & RAdam\\
    Learning rate & - & 0.01\\
    $\beta_0$ & - & 0.5\\
    $\beta_1$ & - & 0.9\\
    Optimizer stride & - & 1\\
    $n_{opt}$ & - & 50\\
    \midrule
    Heuristic function & ArcFace Identity Loss & ArcFace Identity Loss\\
    \bottomrule
    \end{tabularx}
    \label{tab:params}
\end{table}

\clearpage

\subsection{PyTorch Implementation of Greedy Optimization}
\label{appendix:code}
We provide a minimalist example of the greedy optimization procedure for diffusion models in PyTorch.
The code expects noise prediction network \texttt{model(xt,z,t)} that takes the noisy input image \texttt{xt}, conditional \texttt{z}, and timestep \texttt{t} and a scheduler, like the DPM-Solver as seen in \url{https://huggingface.co/docs/diffusers/api/schedulers/multistep_dpm_solver}.
Additionally, the code allows for an optimal parameter \texttt{noise\_level} if the starting sample timestep $t_N < T$ and optimizer striding with \texttt{opt\_stride}.

\begin{algorithm}[!ht]
\begin{minted}[fontsize=\footnotesize,breaklines]{python}
    def greedy_optimization(model, scheduler, xt, z, loss_fn, n_opt_steps=50, opt_stride=1, noise_level=1.0, opt_kwargs={}, device=None):
        device = device if device is not None else model.device

        timesteps = scheduler.timesteps

        if noise_level < 1.0:
            timesteps = timesteps[int((1. - noise_level) * len(timesteps)):]


        for i, t in enumerate(timesteps):
            with torch.no_grad():
                out = model(xt, z, t)

            if (i %
                out = out.detach().clone().requires_grad_(True)
                opt = torch.optim.RAdam([out], **opt_kwargs)

                x0_pred = convert_eps_to_x0(out, t, xt.detach())
                best_loss = loss_fn(x0_pred)
                best_out = out

                for _ in range(n_opt_steps):
                    opt.zero_grad()

                    x0_pred = convert_eps_to_x0(out, t, xt.detach())
                    loss = loss_fn(x0_pred)

                    do_update = (loss < best_loss).float()
                    best_loss = do_update * loss + (1. - loss) * best_loss
                    best_out = do_update[:, None, None, None] * out\
                             + (1. - do_update)[:, None, None, None] * best_out
                             
                    loss.mean().backward()
                    opt.step()

                out = best_out

            xt = scheduler.step(out, t, xt)

        return xt
\end{minted}
\caption{Pytorch Code for Greedy Optimization w.r.t. to a Heuristic function.}
\label{code:greedy_dim}
\end{algorithm}

\section{Evaluation Details}
\label{appendix:eval_details}

\subsection{NFE}
\label{appendix:nfe}
In our reporting of the NFE we record the number of times the diffusion noise prediction U-Net is evaluated both during the encoding phase, $N_E$, and solving of the PF-ODE, $N$.
We chose to report $N +N_E$ over $N + 2N_E$ as even though two bona fide images are encoded resulting in $2N_E$ NFE during encoding, this process can simply be batched together, reducing the NFE down to $N_E$.
When reporting the NFE for the Morph-PIPE model, we report $N_E + BN$ where $B$ is the number of blends.
While a similar argument can be made that the morphed candidates could be generated in a large batch of size $B$, reducing the NFE of the sampling process down to $N$, we chose to report $BN$ as the number of blends, $B = 21$, used in the Morph-PIPE is quite large, potentially resulting in Out Of Memory (OOM) errors, especially if trying to process a mini-batch of morphs.
Using $N_E + N$ reporting over $N_E + BN$, the NFE of Morph-PIPE is 350, which is comparable to DiM.
The NFE for the Greedy-DiM-S algorithm is reported as $N_E + N$ rather than $N_E + 2N$. The justification being that even though we calculate twin trajectories while solving the PF-ODE, this operation can simply be batched, as it only doubles the number of samples.

\subsection{Hardware}
All experiments were done on a single NVIDIA V100 32GB GPU.

\subsection{Datasets}
\label{appendix:datasets}

In the SYN-MAD 2022 dataset the OpenCV morphs only consist of 489 morphed images rather than the full 500 due to technical issues with the remaining 11 morphs.
Due to this we elect to only used the 489 pairs of bona fide images for our experiments to ensure a fair comparison across all morphing attacks.
All the bona fide images from the FRLL dataset are aligned and cropped using the dlib library based on the pre-processing used to align and crop the FFHQ dataset~\cite{stylegan}.
The landmark-based morphs are also aligned in a post-processing step using this strategy.
Since the MIPGAN- and DiM-based morphs already use the alignment script in pre-processing the bona fide images for face morphing, it is unnecessary to run it a second time for these morphing attacks.
All experiments on DiM models used a pre-trained Diffusion Autoencoder trained on the FFHQ dataset as was used in the original papers~\cite{blasingame_dim,morph_pipe,fast_dim}.

\subsection{FR Systems}
\label{appendix:fr_config}
All three FR systems use the Improved ResNet (IResNet-100) architecture~\cite{iresnet} as the neural net backbone for the FR system.
The ArcFace model is widely used FR system~\cite{blasingame_dim,morph_pipe,mipgan,sebastien_gan_threaten}. It employs an additive angular margin loss to enforce intra-class compactness and inter-class distance, which can enhance the discriminative ability of the feature embeddings~\cite{deng2019arcface}.
ElasticFace builds upon the ArcFace model by using an elastic penalty margin over the fixed penalty margin used by ArcFace.
This change results in an FR system with state-of-the-art performance~\cite{elasticface}.
Lastly, the AdaFace model employs an adaptive margin loss by weighting the loss relative to an approximation of the image quality~\cite{adaface}.
The image quality is approximated via feature norms and is used to give less weight to misclassified images, 
reducing the impact of ``low'' quality images on training.
This improvement allows the AdaFace model to achieve state-of-the-art performance in FR tasks.

The AdaFace and ElasticFace models are trained on the MS1M-ArcFace dataset, whereas the ArcFace model is trained on the MS1M-RetinaFace dataset.
\textit{N.B.}, the ArcFace model used in the identity loss is not the same ArcFace model used during evaluation.
The model used in the identity loss is an IResNet-100 trained on the Glint360k dataset~\cite{glint360k_paper} with the ArcFace loss.
We use the cosine distance to measure the distance between embeddings from the FR models.
All three FR systems require images of $112 \times 112$ pixels.
We resize every images, post alignment from dlib which ensures the images are square, to $112 \times 112$ using bilinear down-sampling.
The image tensors are then normalized such that they take value in $[-1, 1]$.
Lastly, the AdaFace FR system was trained on BGR images so the image tensor is shuffled from the RGB format to the BGR format.

\subsection{Detectability Study}
\label{appendix:detect}
Following the approach taken in~\cite{blasingame_dim,fast_dim} we design a detectability study of the morphing attacks using a pre-trained SE-ResNeXt101-32x4d network developed by NVIDIA.
The SE-ResNeXt101-32x4d is a ResNeXt101-32x4d model~\cite{resneXt} with added Squeeze-and-Excitation layers~\cite{squeezenet} pre-trained on the ImageNet~\cite{imagenet} dataset.
We employ $k$-fold stratified cross validation to ensure that the class balance between morphed and bona fide images is preserved across each fold.
We opt to use $k = 5$ for our study.
The powerful SE-ResNeXt101-32x4d  is used as the backbone for the S-MAD detector.
We introduce replace the last layer with a fully connected layer with two outputs which denote the log probabilities of the bona fide and morphed classes.
For each training dataset the backbone is fine-tuned for 3 epochs with an exponential learning rate scheduler and differential learning rates to combat any potential overfitting during training.
We begin with a learning rate of 0.001 on the fully connected layer and reduce the learning rate exponentially to a learning rate of $10^{-7}$ for each layer further away from the fully connected layer.
\textit{N.B.,} the learning rate scheduler has step size of 3 and an exponential rate $\gamma = 0.1$.
To further combat overfitting, the cross entropy loss used for training is used with label smoothing parameter 0.15.
Additionally, we opt to scale the EMA decay $\beta_{ema}$, with the batch size $M$ in accordance to
\begin{equation}
    \beta_{ema} = \bigg(\frac12\bigg)^{\frac{M}{1000}}
\end{equation}
in a manner similar to that of the scaling rule used to update the generator in the Alias-Free GAN~\cite{stylegan3} and recent work has shown the benefit of scaling the EMA decay with batch size~\cite{scale_ema}.
In our experiments we use a batch size $M = 128$, and therefore $\beta_{ema} \approx 0.915$.
\textit{N.B.,} none of the bona fide images and morph pairs used in the training fold are present for the validation fold and the morph pairs in the validation fold are identical for each morphing attack to ensure a fair comparison.

\section{Additional Results}
\label{appendix:addtional_results}

\begin{figure}[h]
    \centering
    \begin{subfigure}{0.49\textwidth}
        \includegraphics[width=\textwidth]{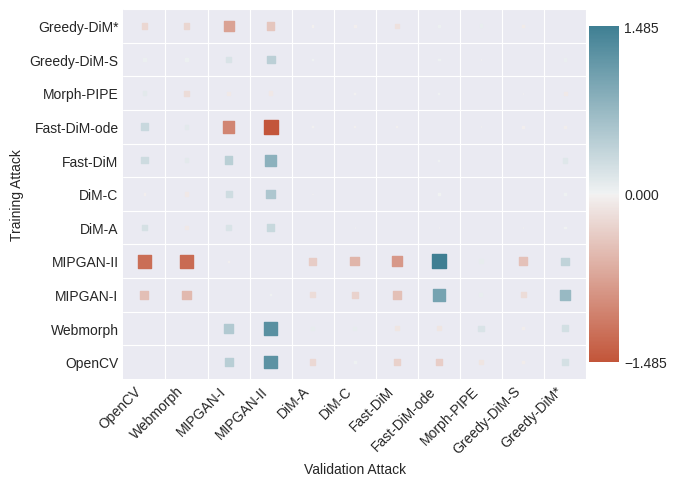}
        \caption{$\Delta(\alpha \| \beta)$ on SYN-MAD 2022.}
       \label{fig:rsm}
    \end{subfigure}
    \begin{subfigure}{0.49\textwidth}
        \includegraphics[width=\textwidth]{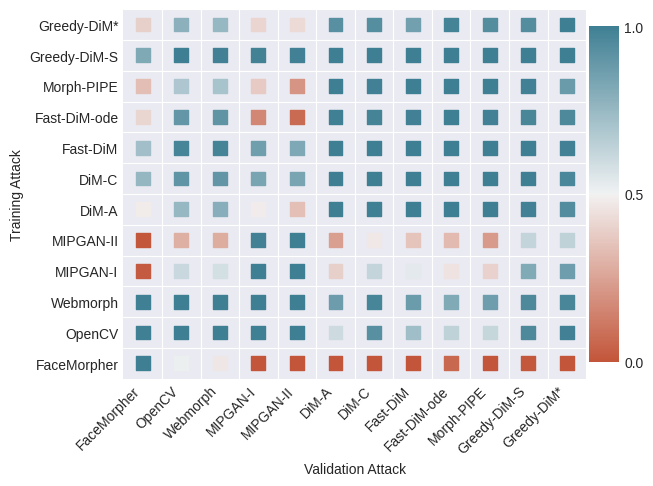}
        \caption{$T(\alpha, \beta)$ on SYN-MAD 2022.}
        \label{fig:transferability}
    \end{subfigure}
    \caption{The RSM and Transferability metrics.}
    \label{fig:rsm_and_t}
\end{figure}

\subsection{Relative Strength Metric}
We also report the Relative Strength Metric (RSM) proposed by Blasingame~\etal~\cite{blasingame_dim}.
The RSM measures how ``strong'' one morphing attack is to another attack, \ie, if a MAD model is trained to detect one attack, how well can it detect the other morphing attack?
For two morphing attacks $\alpha, \beta$, the RSM from $\alpha$ to $\beta$ is defined as
\begin{equation}
\Delta(\alpha \| \beta) = \log \bigg(\frac{T(\alpha, \beta)}{T(\beta, \alpha)}\bigg)
\end{equation}
where $T(\alpha,\beta)$ denotes the probability of a MAD model trained on $\alpha$, $f^\alpha: \X \to \{0,1\}$ is able to detect the morphed attack $\beta$ given $f^\alpha$ detects $\alpha$, \ie,
\begin{equation}
    T(\alpha, \beta) = P(f^\alpha(X^\beta) = 1 \mid f^\alpha(X^\alpha) = 1)
\end{equation}
where $X^\alpha, X^\beta$ denotes morphed images generated by $\alpha$ and $\beta$, respectively.

In~\cref{fig:rsm_and_t} the RSM and transferability metrics are plotted.
Note, in~\cref{fig:rsm} we opt to omit the FaceMorpher entry as it was an outlier with an RSM of approximately $-20$ relative to many of the other morphing attacks, causing scaling issues with the other entries of the plot.
Corroborating our results from~\cref{tab:det_study} the Greedy-DiM* algorithm is slightly ``weaker'' than the other DiM morphs relative to the landmark and GAN-based morphs.
Interestingly we also note that Fast-DiM-ode is ``weaker'' relative to MIPGAN attacks.
This is a strange observation as the only difference between Fast-DiM and Fast-DiM-ode~\cite{fast_dim} is the forward ODE solver and in terms of MMPMR Fast-DiM performs similarly to DiM-C.
We leave further investigation into this matter as future work.
We note that the FaceMorpher attack is especially ``weak'' to other morphs with extremely low transferability to all representation-based morphing attacks, see~\cref{fig:transferability}.
Likewise, the MIPGAN attacks are generally quite ``weak'' as well.

\begin{table}[h]
    \centering
    \caption{Impact of heuristic function on Greedy-DiM*.}
    \footnotesize
    \begin{tabular}{llrrr}
    \toprule
       && \multicolumn{3}{c}{\textbf{MMPMR}($\uparrow$)}\\
            \cmidrule(lr){3-5}
         \textbf{Heuristic} & \textbf{Network} & \textbf{AdaFace} &   \textbf{ArcFace} &  \textbf{ElasticFace}\\
    
    \midrule
     Identity  & ArcFace &               \textbf{99.59} &               98.77 &                   \textbf{99.59} \\
     Worst-Case L2  & ArcFace &              98.77 &               98.16 &                   98.98 \\
     Worst-Case Cosine & ArcFace &              83.84 &               73.21 &                   80.37 \\
     Identity  & LPIPS  &               93.05 &               89.37 &                   93.66 \\
     Worst-Case L2 & LPIPS &               93.25 &               89.78 &                   93.46\\
     Identity + Perceptual Loss & ArcFace and LPIPS &               99.18 &               \textbf{99.18} &                   99.39 \\
    \bottomrule
    \end{tabular}
\label{tab:heuristic}
\end{table}

\subsection{Comparison of all Design Choices}
In this section we presented additional experiments on the impact of different design choices on Greedy-DiM algorithms that were not able to be included in the main paper.
We begin by exploring the impact the choice $\mathcal{H}$ plays on the performance of Greedy-DiM.
We define the worst case loss as the distance between the worst-case morph and the true morph as
\begin{align}
    v_W &= \frac{F(\bfx_0^{(a)}) + F(\bfx_0^{(b)})}{2}\\
    \mathcal{L}_{W} &= d(F(\bfx_0^{(ab)}), v_W)
\end{align}
In this construction $v_W$ is the worst-case morph as it is the morphed code which minimizes L2 distance between both embeddings of the bona fide images.
Alternatively, for the cosine distance the worst-case morph is defined as
\begin{equation}
    v_W = \frac{F(\bfx_0^{(a)}) + F(\bfx_0^{(b)})}{\|F(\bfx_0^{(a)}) + F(\bfx_0^{(b)})\|}
\end{equation}
Recent work~\cite{worst_case_morphs_wasserstein_ali} has explored a similar idea of using the worst-case morphs to guide the generation process of GANs.
We also examine the use of an LPIPS~\cite{LPIPS} trained VGG~\cite{VGG16} network.
In~\cref{tab:heuristic} we outline the impact the choice of heuristic function.
We observe that using a worst-case morph formulation with the ArcFace FR system drops the performance severely over using the identity loss.
Additionally, using the worst-case morph in terms of cosine distance further drops the performance of the Greedy-DiM* algorithm.
We observed that using LPIPS over ArcFace as the neural net backbone with the identity or worst-case losses also decreased performance.
Lastly, we noticed only a marginal change in performance, positive and negative, when incorporating the LPIPS perceptual loss in addition to the ArcFace identity loss.

\begin{table}[h]
    \centering
    \caption{Impact of ODE solver on Greedy-DiM*.}
    \footnotesize
    \begin{tabular}{lrrr}
    \toprule
       & \multicolumn{3}{c}{\textbf{MMPMR}($\uparrow$)}\\
            \cmidrule(lr){2-4}
         \textbf{ODE Solver} & \textbf{AdaFace} &   \textbf{ArcFace} &  \textbf{ElasticFace}\\
    
    \midrule
    DDIM  &               99.59 &               99.18 &                   99.39 \\
    DPM++ 2M &               98.98 &               99.18 &                   99.39 \\
    \bottomrule
    \end{tabular}
\label{tab:todo}
\end{table}

\noindent As suggested in~\cite{fast_dim} we explore using the DPM++ 2M solver proposed by Lu~\etal~\cite{lu2023dpmsolver} over the DDIM solver.
The DPM++ 2M solver is a second-order multi-step ODE solver designed for the PF-ODE.
The drop in slight drop in performance is unsurprisingly, as the multi-step solver requires previous iterations; however, in the Greedy-DiM* algorithm we update the iteration at each step breaking the theoretical assumptions underpinning the derivations in~\cite{lu2023dpmsolver}.
Now as Greedy-DiM* is globally optimal this should lessen the impact, but there is still the variability of gradient descent which could play a role.

\begin{table}[h]
    \centering
    \caption{Impact of initial noise variable.}
    \footnotesize
    \begin{tabular}{llrrr}
    \toprule
       && \multicolumn{3}{c}{\textbf{MMPMR}($\uparrow$)}\\
            \cmidrule(lr){3-5}
         \textbf{Algorithm} & \textbf{Initial }$\bfx_T$ & \textbf{AdaFace} &   \textbf{ArcFace} &  \textbf{ElasticFace}\\
    
    \midrule
    Fast-DiM~\cite{fast_dim} & $\bfx_T = \mathrm{slerp}(\bfx_T^{(a)}, \bfx_T^{(b)}, 0.5)$ & 92.02 & 90.18 & 93.25\\
    Fast-DiM~\cite{fast_dim} & $\bfx_T \sim \mathcal{N}(\mathbf{0}, \mathbf{I})$ & 4.5 & 3.48 & 2.04\\
    \textbf{Greedy-DiM*} & $\bfx_T = \mathrm{slerp}(\bfx_T^{(a)}, \bfx_T^{(b)}, 0.5)$                      &               \textbf{99.59} &               \textbf{98.77} &                   \textbf{99.59} \\
    \textbf{Greedy-DiM*} & $\bfx_T \sim \mathcal{N}(\mathbf{0}, \mathbf{I})$                                      &               70.14 &               75.05 &                   63.8  \\
    \bottomrule
    \end{tabular}
\label{tab:noise}
\end{table}
\noindent Next we examine if it is even necessary to encode the bona fide images into their noisy representations or if we can simply start the Greedy-DiM* algorithm with random noise.
Preechakul~\etal~\cite{diffae} posit that the semantic information is encoded in $\bfz$ and that $\bfx_T$ only encodes the ``stochastic'' information, \ie, information which doesn't impact semantic meaning, like the randomness of hair strands, but not texture or color.
Concurrent work~\cite{fast_dim} has called this assertion into question showing that the stochastic information $\bfx_T$ is essential to creating high quality morphs.
However, since the Greedy-DiM* algorithm searches over the whole image space with a greedy search we explore if it is still necessary to start from the morphed initial noise or if random noise would suffice.
In~\cref{tab:noise} we found the morphed initial noise still plays a large role in helping the gradient descent algorithm find an optimal solution.
We did notice, however, that the Greedy-DiM* algorithm fared much better than the Fast-DiM algorithm when starting with random noise resulting in a roughly 70\% increase in MMPMR.

\begin{table}[h]
    \centering
    \caption{Impact of the number of sampling steps on Greedy-DiM*.}
    \footnotesize
    \begin{tabular}{lrrr}
    \toprule
       & \multicolumn{3}{c}{\textbf{MMPMR}($\uparrow$)}\\
            \cmidrule(lr){2-4}
         $N$ & \textbf{AdaFace} &   \textbf{ArcFace} &  \textbf{ElasticFace}\\
         \midrule
    
     20                           &               99.59 &               98.77 &                   99.59 \\
     50         &               99.18 &               99.18 &                   99.39 \\
    100                       &               99.59 &               99.18 &                   99.39 \\
    \bottomrule
    \end{tabular}
\label{tab:n_steps}
\end{table}

\noindent We perform a small study on the impact of the number of sampling steps on the MMPMR of the resulting morphed image the results of which are shown in~\cref{tab:n_steps}.
We observe that the number of sampling steps has a negligible impact on performance and so we opt to use $N = 20$ to conserve compute resources.

\begin{table}[h]
    \centering
    \caption{Impact of optimizer momentum parameters on Greedy-DiM*.}
    \footnotesize
    \begin{tabular}{llrrr}
    \toprule
       && \multicolumn{3}{c}{\textbf{MMPMR}($\uparrow$)}\\
            \cmidrule(lr){3-5}
         $\beta_0$ & $\beta_1$ & \textbf{AdaFace} &   \textbf{ArcFace} &  \textbf{ElasticFace}\\
    
    \midrule
     0.9 & 0.999                      &               99.59 &               98.77 &                   99.59 \\
     0.5 & 0.9              &              100    &              100    &                  100    \\
    \bottomrule
    \end{tabular}
\label{tab:betas}
\end{table}

\noindent Lastly, in~\cref{tab:betas} we explore the impact of the optimizer momentum parameters on the Greedy-DiM* algorithm.
The Adam~\cite{Kingma2015AdamAM} optimizer, and it's derivatives like RAdam~\cite{liu2019radam}, recommend the use of the parameters $\beta_0 = 0.9$ and $\beta_1 = 0.999$ by default.
We explore alternative parameters that were used by Gu~\etal~\cite{weird_betas} with $\beta_0 = 0.5$ and $\beta_1 = 0.9$.
We find that these parameters are the final missing piece in maximizing the MMPMR performance on the studied FR systems resulting in a 100\% MMMPR at FMR = 0.1\%.
We posit that because the parameters, $\beta_0, \beta_1$ control the exponential decay rates of the Adam optimizer, while the initial recommended parameter settings are appropriate for training a neural network or gradient descent on, the $\boldsymbol\epsilon$ space needs a faster update to converge quickly, resulting in the lower decay rates.
Moreover, as our search space is considerably simpler than that of the parameter space of a large neural network, we can reduce the amount of smoothing used by the optimizer.

\begin{table}[h]
\caption{Time needed to create 500 morphed images on a single NVIDIA Tesla V100 32GB PCIe GPU. \textdagger~denotes the batched implementation of Morph-PIPE.}
\centering
\footnotesize
\begin{tabularx}{\linewidth}{@{\extracolsep{\fill}}lrrr}
\toprule
\textbf{Model} & \textbf{NFE} & \textbf{Batch Size} & \textbf{Time (HH:MM:SS)} \\
\midrule
Fast-DiM-ode~\cite{fast_dim} & 150 & 32 & 01:10:14 \\
Fast-DiM~\cite{fast_dim} & 300 &48 & 01:30:01 \\
DiM-C~\cite{blasingame_dim} & 350 & 48 & 01:43:16 \\
DiM-A~\cite{blasingame_dim} & 350 & 48 & 01:43:17 \\
\textbf{Greedy-DiM*} & 270 & 40 & 02:28:57 \\
\textbf{Greedy-DiM-S} & 350 & 48 & 02:54:47 \\
Morph-PIPE~\cite{morph_pipe} & 2350 & 48 & 07:25:16 \\
Morph-PIPE\textsuperscript{\textdagger}~\cite{morph_pipe} & 350 & 4 & 07:35:22 \\
\bottomrule
\end{tabularx}
\label{tab:speed}
\end{table}

\subsection{Empirical Time Analysis}
While we chose to report NFE as our measure of efficiency as it is a standard method of reporting the computational cost of diffusion models and is invariant to the underlying hardware, for completeness we report the time to generate the morphed images used in this paper using our hardware.
We recorded the time it took to generate all 500 morphed images from the SYN-MAD 2022 dataset and reported our results in~\cref{tab:speed}.
For each model we selected the largest batch size that would fit on our GPU without running into Out Of Memory errors during the creation of the morphs.
\textit{N.B.}, for increased efficiency of GEMM kernel calls we used an integer multiple of 8 whenever possible.
\cref{tab:speed} shows that while the search and optimization step of the Greedy-DiM algorithm adds some overhead compared to the original DiM morphs, or especially the recently proposed Fast-DiM morphs, in comparison with the only other identity guided DiM model, Morph-PIPE, our added overhead is far less.
Combined with the superior performance of Greedy-DiM*, we believe that this represents a significant contribution as it is far faster than Morph-PIPE with a small amount of overhead compared to the original DiM morphs.

For completeness, we also implement an alternative implementation for Morph-PIPE, wherein the candidate morphs are generated in a batch of size 21 blends instead of being calculated sequentially; however, due to the computationally demanding requirements of generating all 21 blends at once, we had to significantly reduce the batch size in order to fit into the memory of our GPU.
We denote this implementation as Morph-PIPE\textsuperscript{\textdagger} in~\cref{tab:speed}.
This approach is therefore limited only to groups with significant amounts of computing resources.
Conversely, our method can be used by groups with more modest hardware, providing greater accessibility to researchers.
Interestingly, the batched implementation of Morph-PIPE takes slightly longer than the regular implementation, however, this difference is quite small and unlikely to be significant.
We believe the experimental results from~\cref{tab:speed} justify the reasoning used to make our choices in~\cref{appendix:nfe}.

The overhead introduced by Greedy-DiM is rather minimal, as the search/optimization step of the Greedy-DiM models is \textit{relatively} cheap as it involves only performing a search over the possible $\boldsymbol\epsilon = \boldsymbol\epsilon_\theta(\bfx_t, \bfz, t)$.
As such, the gradient calculation is fairly cheap, as we do \textit{not} backpropagate through the U-Net $\boldsymbol\epsilon_\theta$. It can be shown that the gradient of the heuristic function w.r.t. $\boldsymbol\epsilon$ involves only a single call of an automatic differentiation algorithm, apart from any additional calls to backpropogate through the heuristic function, to evaluate $\nabla_\bfx\mathcal{H}$.
Remember that $\hat{\bfx}_0 = \frac{\bfx_{t} - \sigma_t\boldsymbol\epsilon}{\alpha_t}$
then the gradient of interest is
$\nabla_{\boldsymbol\epsilon} \mathcal{H} = - \frac{\sigma_t}{\alpha_t} \nabla_{\bfx} \mathcal{H}$
which is only as expensive as the image size and heuristic function.

\clearpage
\section{Additional Images}
\label{appendix:additional_images}
\begin{figure}[h]
    \centering
    \includegraphics[width=0.245\textwidth]{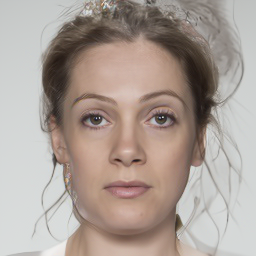}
    \includegraphics[width=0.245\textwidth]{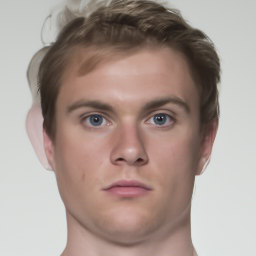}
    \includegraphics[width=0.245\textwidth]{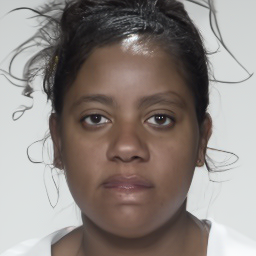}
    \includegraphics[width=0.245\textwidth]{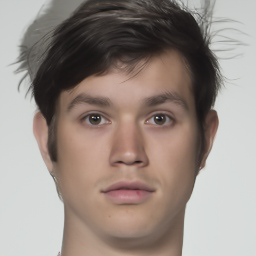}

    \includegraphics[width=0.245\textwidth]{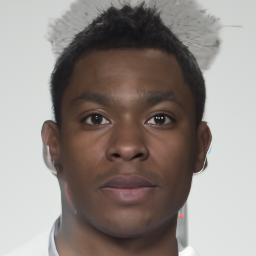}
    \includegraphics[width=0.245\textwidth]{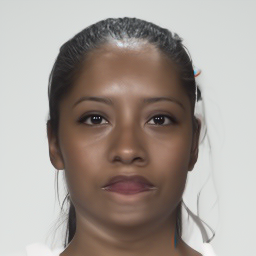}
    \includegraphics[width=0.245\textwidth]{figures/frll/greedy_dim_s/096_137.png}
    \includegraphics[width=0.245\textwidth]{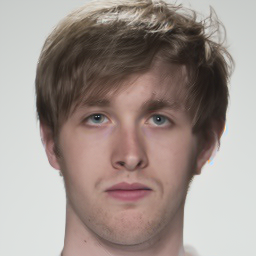}
    \caption{Morphed images generated via Greedy-DiM-S.}
    \label{fig:greedy-dim-s-morphs}
\end{figure}

\begin{figure}[h]
    \centering
    \includegraphics[width=0.245\textwidth]{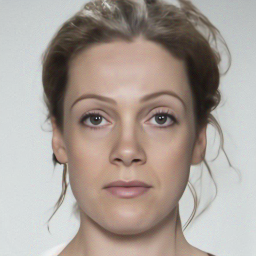}
    \includegraphics[width=0.245\textwidth]{figures/frll/greedy_dim_star/012_138.png}
    \includegraphics[width=0.245\textwidth]{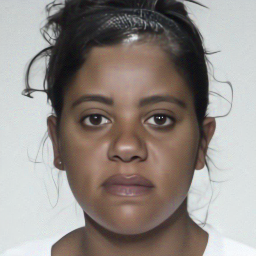}
    \includegraphics[width=0.245\textwidth]{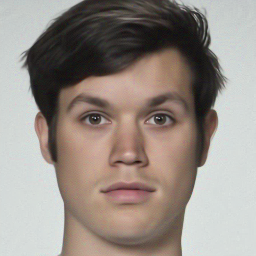}

    \includegraphics[width=0.245\textwidth]{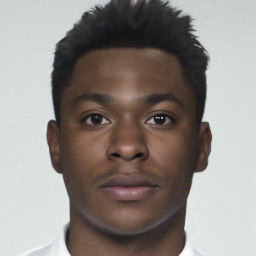}
    \includegraphics[width=0.245\textwidth]{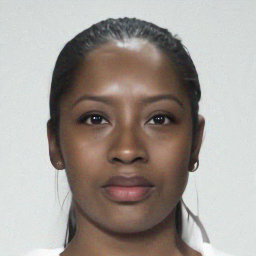}
    \includegraphics[width=0.245\textwidth]{figures/frll/greedy_dim_star/096_137.png}
    \includegraphics[width=0.245\textwidth]{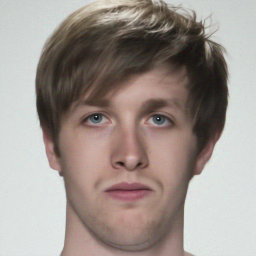}
    \caption{Morphed images generated via Greedy-DiM*.}
    \label{fig:greedy-dim-star-morphs}
\end{figure}

In this section we provide additional samples produced by the Greed-DiM-S algorithm,~\cref{fig:greedy-dim-s-morphs}, and Greedy-DiM* algorithm,~\cref{fig:greedy-dim-star-morphs}.
In~\cref{fig:greedy-dim-star-morphs} we observe consistent high frequency noise artefacts that looks akin to film grain whereas the the Greedy-DiM-S, and other DiM variants for that matter, lack this quality.
We posit this a by product of the optimization process and this noise encodes information to trick the FR system similar to the technique of adversarial perturbations~\cite{adversarial_perturbations}.

\end{document}